\documentclass[preprint,12pt]{elsarticle}

\usepackage{hyperref}
\usepackage{url}
\usepackage{amsmath}
\usepackage{amssymb}
\usepackage{amsfonts}
\usepackage{leftidx}

\usepackage{amsthm}
\usepackage{bm}
\usepackage{bbm}
\usepackage[labelformat=simple]{subcaption}

\usepackage[normalem]{ulem}
\usepackage{enumitem}
\usepackage[ruled,linesnumbered]{algorithm2e}
\usepackage{multicol}
\usepackage{xcolor,colortbl}

\definecolor{Gray}{gray}{0.85}
\definecolor{LightCyan}{rgb}{0.88,1,1}

\newcommand\ci{\perp\!\!\!\perp}

\def\presuper#1#2%
  {\mathop{}%
   \mathopen{\vphantom{#2}}^{#1}%
   \kern-\scriptspace%
   #2}
   
   \makeatletter
\newcommand*\rel@kern[1]{\kern#1\dimexpr\macc@kerna}
\newcommand*\widebar[1]{%
  \begingroup
  \def\mathaccent##1##2{%
    \rel@kern{0.8}%
    \overline{\rel@kern{-0.8}\macc@nucleus\rel@kern{0.2}}%
    \rel@kern{-0.2}%
  }%
  \macc@depth\@ne
  \let\math@bgroup\@empty \let\math@egroup\macc@set@skewchar
  \mathsurround\z@ \frozen@everymath{\mathgroup\macc@group\relax}%
  \macc@set@skewchar\relax
  \let\mathaccentV\macc@nested@a
  \macc@nested@a\relax111{#1}%
  \endgroup
}
\makeatother

\newtheorem{definition1}{Definition}
\newtheorem{theorem1}{Theorem}
\newtheorem{lemma1}{Lemma}
\newtheorem{proposition1}{Proposition}

\newtheorem{listn}{List}

\usepackage{tikz}
\usepackage{tensor}
\usetikzlibrary{decorations.pathreplacing,angles,quotes}
\usepackage{tkz-tab}
\usetikzlibrary{plotmarks}

\usetikzlibrary{arrows, automata}
\usetikzlibrary{decorations.pathreplacing,angles,quotes}

\def\circarrow{{\circ\hspace{0.3mm}\!\!\! \rightarrow}}
\def\circlinecirc{{\circ \hspace{0.4mm}\!\!\! - \hspace{0.4mm}\!\!\!\circ}}
\def\circline{{\circ \! -}}
\def\linecirc{{- \! \circ}}

\usepackage[margin=1.15in]{geometry}

\makeatletter
\newtheorem*{rep@theorem}{\rep@title}
\newcommand{\newreptheorem}[2]{%
\newenvironment{rep#1}[1]{%
 \def\rep@title{#2 \ref{##1}}%
 \begin{rep@theorem}}%
 {\end{rep@theorem}}}
\makeatother

\newreptheorem{theorem}{Theorem}
\newreptheorem{lemma}{Lemma}
\newreptheorem{proposition}{Proposition}

\begin{document}
\begin{frontmatter}
\title{Causal Discovery with a Mixture of DAGs}

\author{Eric V. Strobl}
\address{Psychiatry \& Behavioral Sciences\\Vanderbilt University Medical Center\\Tennessee, United States}

\begin{abstract}Causal processes in biomedicine may contain cycles, evolve over time or differ between populations. However, many graphical models cannot accommodate these conditions. We propose to model causation using a mixture of directed cyclic graphs (DAGs), where the joint distribution in a population follows a DAG at any single point in time but potentially different DAGs across time. We also introduce an algorithm called Causal Inference over Mixtures that uses longitudinal data to infer a graph summarizing the causal relations generated from a mixture of DAGs. Experiments demonstrate improved performance compared to prior approaches.
\end{abstract}
\end{frontmatter}

\section{Introduction}

Causal discovery refers to the process of inferring causation from data. Investigators usually perform causal discovery in biomedicine using randomized controlled trials (RCTs). However, RCTs can be impractical or unethical to perform. For example, scientists cannot randomly administer illicit substances or traumatize healthy subjects. Many investigators therefore experiment with animals knowing that the derived results may not directly apply to humans.

In this paper, we develop an algorithm that discovers causation directly from human observational data, or data collected without randomization. Denote the variables in an observational dataset by $\bm{X}$. We summarize the causal relations between variables in $\bm{X}$ using a \textit{directed graph}, where the directed edge $X_i \rightarrow X_j$ with $X_i,X_j \in \bm{X}$ means that $X_i$ is a \textit{direct cause} of $X_j$. Similarly, $X_i$ is a \textit{cause} of $X_j$ if there exists a directed path, or a sequence of directed edges, from $X_i$ to $X_j$. We want to recover the directed graph as best as possible using the observational dataset.

Directed graphs in nature often contain feedback loops, or \textit{cycles}, where $X_i$ causes $X_j$ and $X_j$ directly causes $X_i$. For example, Figure \ref{fig_cycle} depicts a portion of the thyroid system where $X_1$ denotes the thyroid stimulating hormone (TSH) and $X_2$ the T4 hormone (T4). TSH released from the thyroid gland regulates T4 hormone release, while T4 feeds back to inhibit TSH release. Cycles such as these abound in biomedicine, so we must develop algorithms that can accommodate them in order to accurately model causal processes.

We propose to model a potentially cyclic causal process using multiple directed acyclic graphs (DAGs), or graphs with directed edges but no cycles. The causal process is represented as a DAG at any single point in time, but the DAG may change across time to accommodate feedback. We illustrate the idea by decomposing the cycle in Figure \ref{fig_cycle} into two DAGs: TSH $\rightarrow$ T4 and T4 $\rightarrow$ TSH. For each sample, TSH first causes T4 release at time point $t_1$ and then T4 inhibits TSH release at time point $t_2>t_1$. We however can only measure each sample at a single point in time, so the observational dataset in Figure \ref{table_SB1} contains some samples in blue when TSH causes T4 and others in grey when T4 causes TSH. If we do not observe the time variable $T$, then the observational dataset arises from a \textit{mixture of DAGs} where the mixing occurs over time: $f(X_1,X_2) = \sum_T f(X_1,X_2|T)f(T)$. We must infer the directed graph in Figure \ref{fig_cycle} using the samples from $X_1$ and $X_2$ alone. In practice, we observe more than two random variables without color coding and mixing occurs over a subset of variables $\bm{T}$ denoting entities such as time, gender, income and disease status. Figure \ref{table_SB2} therefore depicts a more realistic dataset.

\begin{figure*}
\centering
\begin{subfigure}{0.15\textwidth}
\centering
\resizebox{\linewidth}{!}{
\begin{tikzpicture}[scale=1.0, shorten >=1pt,auto,node distance=2.8cm, semithick]
                    
\tikzset{vertex/.style = {inner sep=0.4pt}}
\tikzset{edge/.style = {->,> = latex'}}
 
\node[vertex] (1) at  (0,0) {$X_1$};
\node[vertex] (2) at  (1.5,0) {$X_2$};

\draw[edge, bend right] (1) to (2);
\draw[edge, bend right] (2) to (1);
\end{tikzpicture}
}
\caption{}  \label{fig_cycle}
\end{subfigure}
\vspace{5mm}

\begin{subtable}{0.45\textwidth}
  \centering
\begin{tabular}{c c c}
  $X_1$ & $X_2$ & $T$ \\
	\hline
  \rowcolor{LightCyan}0.21 & -0.20 & 1.29 \\
  \rowcolor{Gray}0.68 & -0.47 & 7.30 \\
  \rowcolor{LightCyan}1.05 & -0.19 & 4.33 \\
  \rowcolor{LightCyan}0.72 & -1.40 & 0.10 \\
  \rowcolor{Gray}0.13 & -0.56 & 2.91 \\ 	
  \hline
\end{tabular}
\caption{} \label{table_SB1}
\end{subtable}
\begin{subtable}{0.45\textwidth}
  \centering
\begin{tabular}{ c c c c c c}
  $X_1$ & $X_3$ & $X_4$ & $X_7$ & $\cdots$ & $T_3$\\
	\hline
  0.31 & -1.01 & 5 & 0 & $\cdots$ & 1.29\\ 
  0.89 & -0.58 & 6 & 0 & $\cdots$ & 7.30\\
  1.11 & -0.79 & 2 & 1 & $\cdots$ & 4.33\\
  0.14 & -1.23 & 5 & 0 & $\cdots$ & 0.10\\
  0.21 & -0.20 & 4 & 1 & $\cdots$ & 2.91\\	
  $\vdots$ & $\vdots$ & $\vdots$ & $\vdots$ & $\vdots$ & $\vdots$\\
  \hline
\end{tabular}
\caption{} \label{table_SB2}
\end{subtable}
\caption{We decompose the cycle in (a) into two DAGs: $X_1 \rightarrow X_2$ and $X_2 \rightarrow X_1$. The blue samples in (b) refer to samples arising from the first DAG and the grey ones to the second. The table in (c) depicts a more realistic dataset containing more variables and samples. } \label{fig_cycle_all}
\end{figure*}

We also develop a method for recovering a directed graph summarizing the causal relations arising from a mixture of DAGs. We do so by first reviewing related work in Section \ref{sec_past}. We then provide background in Section \ref{sec_DAGs}. Section \ref{sec_mix} introduces the mixture of DAGs framework. In Section \ref{sec_algo}, we detail the algorithm called Causal Inference over Mixtures (CIM) to infer causal relations using longitudinal data. We then report experimental results in Section \ref{sec_exp} highlighting the superiority of CIM compared to prior approaches on both real and synthetic datasets. We finally conclude the paper in Section \ref{sec_conc}. We delegate all proofs to the Appendix.

This paper improves upon a previous conference paper \citep{Strobl19a}. We made the following changes: (1) simplified exposition, (2) improved characterization of a mixture of DAGs, (3) corrected theoretical results, (4) an enhanced CIM algorithm, (5) experiments with better evaluation metrics. This report therefore provides more convincing material compared to the conference paper.

\section{Related Work} \label{sec_past}

Several algorithms perform causal discovery with cycles. Most of these methods assume  \textit{stationarity}, or a stable distribution over time and populations. The Fast Causal Inference (FCI) algorithm for example infers causal relations when cycles exist \citep{Spirtes00,Zhang08}. The algorithm was initially developed for the acyclic case, but it can infer the acyclic portions of a cyclic graph by ignoring the independence relations within cycles. Other algorithms attempt to recover within-cycle causal relations. The Cyclic Causal Discovery (CCD) algorithm for instance works well when no selection bias or latent variables exist. The Cyclic Causal Inference (CCI) algorithm extends CCD to handle selection bias and latent variables, but both algorithms require linear or discrete variables for correctness \citep{Strobl18,Forre17,Forre18}.

Investigators have extended FCI, CCD and CCI with answer set programming (ASP). ASP algorithms allow the user to easily incorporate prior knowledge and infer causal relations more accurately. These methods however only apply to datasets with less than 10-20 variables due to scalability issues with a conventional laptop \citep{Hyttinen13,Hyttinen14}.

Another set of methods focus on non-stationarity, but most of them require a single underlying directed graph \citep{Dagum95,Blondel17,Zhang17,Magliacane16,Triantafillou15}. Two methods exist for recovering causal processes with multiple graphs \citep{Stroblthesis17,Zhang18}, but they assume a mixture of parametric distributions. CIM improves upon all of these methods by allowing non-linearity, cycles, non-stationarity, non-parametric distributions, changing graphical structure, latent variables and selection bias. 

\section{Background} \label{sec_DAGs}

We now delve into the background material required to understand the proposed methodology. 
\subsection{Terminology}

In addition to directed edges, we consider other edge types including: $\leftrightarrow$ (bidirected), --- (undirected), $\circarrow$ (partially directed), $\circline$ (partially undirected) and $\circlinecirc$ (nondirected). The edges contain three endpoint types: arrowheads, tails and circles. We say that two vertices $X_i$ and $X_j$ are \textit{adjacent} if there exists an edge between the two vertices. We refer to the triple $X_i *\!\! \rightarrow X_j \leftarrow \!\! * X_k$ as a \textit{collider} or \textit{v-structure}, where each asterisk corresponds to an arbitrary endpoint type. A collider or v-structure is said to be \textit{unshielded} when $X_i$ and $X_k$ are non-adjacent. The triple $X_i *\!\! - \!\!* X_j *\!\! - \!\!* X_k$ is conversely a \textit{triangle} if $X_i$ and $X_k$ are adjacent. Unless stated otherwise, a \textit{path} is a sequence of edges without repeated vertices. $X_i$ is an \textit{ancestor} of $X_j$ if there exists a directed path from $X_i$ to $X_j$ or $X_i=X_j$. We write $X_i \in \textnormal{Anc}_{\mathbb{G}}(X_j)$ when $X_i$ is an ancestor of $X_j$ in the graph $\mathbb{G}$. We also apply the definition of an ancestor to a set of vertices $\bm{Y} \subseteq \bm{X}$ as follows: 
\begin{equation} \nonumber
\begin{aligned}
\textnormal{Anc}_{\mathbb{G}}(\bm{Y}) &= \{X_i | X_i \in \textnormal{Anc}_{\mathbb{G}}(Y_j) \text{ for some } Y_j \in \bm{Y}\}.
\end{aligned}
\end{equation}

If $\bm{A}$, $\bm{B}$ and $\bm{C}$ are disjoint sets of vertices in $\bm{X}$, then $\bm{A}$ and $\bm{B}$ are said to be \textit{d-connected} by $\bm{C}$ in a directed graph $\mathbb{G}$ if there exists a path $\Pi$ between some vertex in $\bm{A}$ and some vertex in $\bm{B}$ such that, for any collider $X_i$ on $\Pi$, $X_i$ is an ancestor of $\bm{C}$ and no non-collider on $\Pi$ is in $\bm{C}$. We also say that $\bm{A}$ and $\bm{B}$ are \textit{d-separated} by $\bm{C}$ if they are not d-connected by $\bm{C}$. For shorthand, we write $\bm{A} \ci_d \bm{B} | \bm{C}$ to denote d-separation and $\bm{A} \not \ci_d \bm{B} | \bm{C}$ to denote d-connection. The set $\bm{C}$ is more specifically called a \textit{minimal separating set} if we have $\bm{A} \ci_d \bm{B} | \bm{C}$ but $\bm{A} \not \ci_d \bm{B} | \bm{D}$, where $\bm{D}$ denotes any proper subset of $\bm{C}$.

A \textit{mixed graph} contains edges with only arrowheads or tails, while a \textit{partially oriented mixed graph} may also include circles. We focus on mixed graphs that contain at most one edge between any two vertices. We can associate a mixed graph $\mathbb{G}^*$ with a directed graph $\mathbb{G}$ as follows. We first partition $\bm{X} = \bm{O} \cup \bm{L} \cup \bm{S}$ denoting observed, latent and selection variables, respectively. We then consider a graph over $\bm{O}$ summarizing the ancestral relations in $\mathbb{G}$ with the following endpoint interpretations: $O_i * \!\! \rightarrow O_j$ in $\mathbb{G}^*$ if $O_j \not \in \textnormal{Anc}_{\mathbb{G}}(O_i \cup \bm{S})$, and $O_i * \!\! \textnormal{---} O_j$ in $\mathbb{G}^*$  if $O_j \in \textnormal{Anc}_{\mathbb{G}}(O_i \cup \bm{S})$.

\subsection{Probabilistic Interpretation}
We associate a density $f(\bm{X})$ to a DAG $\mathbb{G}$ by requiring that the density factorize into the product of conditional densities of each variable given its parents:
\begin{equation} \label{eq_fac}
f(\bm{X})=\prod_{i=1}^{p} f(X_i | \textnormal{Pa}_{\mathbb{G}}(X_i)).
\end{equation}
 Any distribution which factorizes as above also satisfies the \textit{global Markov property} w.r.t. $\mathbb{G}$ where, if we have $\bm{A} \ci_d \bm{B} | \bm{C}$ in $\mathbb{G}$, then $\bm{A}$ and $\bm{B}$ are conditionally independent given $\bm{C}$ \citep{Lauritzen90}. We denote the conditional independence (CI) as $\bm{A} \ci \bm{B} | \bm{C}$ for short. We refer to the converse of the global Markov property as \textit{d-separation faithfulness}. An algorithm is \textit{constraint-based} if it utilizes CI testing to recover some aspects of $\mathbb{G}^*$ as a consequence of the global Markov property and d-separation faithfulness.

\section{Mixture of DAGs} \label{sec_mix}
We introduce the framework with univariate $T$ and then generalize to multivariate $\bm{T}$ because the univariate case is simpler.

\subsection{Univariate Case}
We consider the set of vertices $\bm{Z}=\bm{X}\cup T$. We divide $\bm{Z}$ into three non-overlapping sets $\bm{O}$, $\bm{L}$ and $\bm{S}$ denoting observed, latent and selection variables, respectively. At each time point $t$, we consider the joint density $f(\bm{X},T=t)$ and assume that it factorizes according to a DAG $\mathbb{G}^t$ over $\bm{Z}$: 
\begin{align}
f(\bm{X},T=t) &= f(T=t)f(\bm{X}|T=t) \nonumber\\
&= f(T=t)\prod_{i=1}^p f(X_i|\textnormal{Pa}_t(X_i)), \nonumber
\end{align}
where $\textnormal{Pa}_t(Z_i)$ refers to $\textnormal{Pa}_{\mathbb{G}^t}(Z_i)$ for shorthand, the parent set of $Z_i$ at time point $t$. We analyze the following density:
\begin{align}
f(\bm{Z}) &= \prod_{i=1}^{p+1} f(Z_i|\textnormal{Pa}_T(Z_i)), \label{eq_fac_1T}
\end{align}
where $\textnormal{Pa}_T(T) = \emptyset$. The above equation differs from Equation \eqref{eq_fac} for a single DAG; the parent set $\textnormal{Pa}(Z_i)$ remains constant over time in Equation \eqref{eq_fac}, but the parent set $\textnormal{Pa}_T(Z_i)$ may vary over time in Equation \eqref{eq_fac_1T}.

Let $\bm{R} \subseteq \bm{Z}$ correspond to all those variables in $\bm{Z}$ where $T \not \in \textnormal{Pa}_T(Z_i)$, so that $T \in \textnormal{Pa}_T(Z_i)$ for all $Z_i \in [\bm{Z} \setminus \bm{R}]$. We can then rewrite Equation \eqref{eq_fac_1T}:
\begin{align}
&\hspace{3mm}\prod_{i=1}^{p+1} f(Z_i|\textnormal{Pa}_T(Z_i)) \nonumber\\
=&\prod_{Z_i \in \bm{R}} f(Z_i|\textnormal{Pa}_{T}(Z_i) \setminus T)\prod_{Z_i \in [\bm{Z} \setminus \bm{R}]} f(Z_i|\textnormal{Pa}_T(Z_i)\cup T). \label{eq_fac_endT}
\end{align}
The left hand term corresponds to the stationary component and the right hand to the non-stationary component. We assume that we can sample from the mixture density $f(\bm{O}|\bm{S})$:
\begin{equation}
\begin{aligned}
f(\bm{O}|\bm{S}) &= \sum_{\bm{L}} f(\bm{O},\bm{L}|\bm{S}), \nonumber
\end{aligned}
\end{equation}
where mixing occurs over time $T$ in the integration if $T \in \bm{L}$. We refer to the above equation as the \textit{mixture of DAGs} framework. 

\subsection{Multivariate Case}

We generalize the mixture of DAGs framework to a multivariate set of mutually independent \textit{mixture variables} $\bm{T}$. For example, we may let $\bm{T} = \{T_1, T_2\}$, where $T_1$ denotes time and $T_2$ gender. Gender is instantiated independent of time, but the causal process can change over time and differ by gender. We may also observe gender but not observe time so that $T_2 \in \bm{O}$ but $T_1 \in \bm{L}$. The set $\bm{T}$ can therefore encompass a wide range of variables.

We consider the set of vertices $\bm{Z} = \bm{X} \cup \bm{T}$ instead of the original $\bm{X}\cup T$. We divide $\bm{Z}$ into three non-overlapping sets $\bm{O}$, $\bm{L}$ and $\bm{S}$. We assume a joint density $f(\bm{X},\bm{T})$ that factorizes according to a DAG $\mathbb{G}^{\bm{T}}$ over $\bm{Z}$:
\begin{align}
f(\bm{Z}) &= f(\bm{T})f(\bm{X}|\bm{T}) = \prod_{i=1}^s f(T_i)\prod_{i=1}^{p} f(X_i|\textnormal{Pa}_{\bm{T}}(X_i)) \nonumber\\
& = \prod_{i=1}^{p+s} f(Z_i|\textnormal{Pa}_{\bm{T}}(Z_i)),\label{eq_fac_1M}
\end{align}
where $\textnormal{Pa}_{\bm{T}}(\bm{T}) = \emptyset$. The above equation mirrors Equation \eqref{eq_fac_1T}.

For each $Z_i \in \bm{Z}$, let $\bm{U}_i \subseteq \bm{T}$ denote the largest set such that $\bm{U}_i \cap \textnormal{Pa}_{\bm{T}}(Z_i) = \emptyset$. This implies $\bm{T} \cap \textnormal{Pa}_{\bm{T}}(Z_i) = \bm{T} \setminus \bm{U}_i \triangleq \bm{V}_i$. We then rewrite Equation \eqref{eq_fac_1M}:
\begin{align}
&\prod_{i=1}^{p+s}f(Z_i|\textnormal{Pa}_{\bm{T}}(Z_i)) =\prod_{i=1}^{p+s} f(Z_i|(\textnormal{Pa}_{\bm{T}}(Z_i) \setminus \bm{U}_i) \cup \bm{V}_i), \label{eq_fac_endM}
\end{align}
so that $f(Z_i|\textnormal{Pa}_{\bm{T}}(Z_i))$ is stationary over $\bm{U}_i$ but non-stationary over $\bm{V}_i$. Setting $\bm{U}_i = T$ and $\bm{V}_i = \emptyset$ for $Z_i \in \bm{R}$ and vice versa for $Z_i \in [\bm{Z} \setminus \bm{R}]$ recovers Equation \eqref{eq_fac_endT}. We finally sample from the mixture density $f(\bm{O}|\bm{S})$:
\begin{equation} \nonumber
f(\bm{O}|\bm{S}) = \sum_{\bm{L}} f(\bm{O},\bm{L}|\bm{S}).
\end{equation}

\subsection{Global Markov Property} \label{sec_markov}

The factorization in Equation \eqref{eq_fac_endM} implies certain CI relations. In this section, we will identify the CI relations by deriving a global Markov property similar to the traditional DAG case. 

There exists a DAG $\mathbb{G}_{\bm{T}}$ for each instantiation of $\bm{T}$ because $\textnormal{Pa}_{\bm{T}}(Z_i)$ is defined for all $Z_i \in \bm{Z}$. Consider the collection $\mathcal{G}$ consisting of all DAGs indexed by $\bm{T}$. The number of DAGs over $\bm{Z}$ is finite, so $|\mathcal{G}|=q \in \mathbb{N}^+$. Let $\mathcal{T}$ denote the values of $\bm{T}$ corresponding to any member of $\mathcal{G}$, and $\mathcal{T}^j$ to those for $\mathbb{G}^j \in \mathcal{G}$. We can then rewrite Equation \eqref{eq_fac_endM} as:
\begin{align} \nonumber
&\prod_{i=1}^{p+s}f(Z_i|\textnormal{Pa}_{\bm{T}}(Z_i)) =\sum_{j=1}^q \mathbbm{1}_{\bm{T} \in \mathcal{T}^j} \prod_{i=1}^{p+s} f(Z_i|\textnormal{Pa}^j(Z_i)),
\end{align}
where $\textnormal{Pa}^j(Z_i)$ refers to the parents of $Z_i$ in $\mathbb{G}^j$, and $\textnormal{Ch}^j(Z_i)$ the children. If $\bm{A} \subseteq \bm{Z}$, then we use $\bm{A}^j$ to more directly refer to the vertices corresponding to $\mathbb{G}^j$. We let $\bm{A}^\prime = \cup_{j=1}^q \bm{A}^j$.

\begin{figure*}
\centering
\begin{subfigure}{0.40\textwidth}
\centering
\resizebox{\linewidth}{!}{
\begin{tikzpicture}[scale=1.0, shorten >=1pt,auto,node distance=2.8cm, semithick]
                    
\tikzset{vertex/.style = {inner sep=0.4pt}}
\tikzset{edge/.style = {->,> = latex'}}
 
\node[vertex] (1) at  (0,0) {$X_1^1$};
\node[vertex] (3) at  (1.5,0) {$X_3^1$};
\node[vertex] (2) at  (0.75,-1.5) {$X_2^1$};
\node[vertex] (7) at  (2.25,-1.5) {$T_1^1$};

\draw[edge] (1) to (2);

\node[vertex] (4) at  (4,0) {$X_3^2$};
\node[vertex] (6) at  (5.5,0) {$X_1^2$};
\node[vertex] (5) at  (4.75,-1.5) {$X_2^2$};
\node[vertex] (8) at  (3.25,-1.5) {$T_1^2$};

\draw[edge] (5) to (4);
\draw[edge] (7) to (2);
\draw[edge] (8) to (5);
\draw[edge] (7) to (3);
\draw[edge] (8) to (4);
\end{tikzpicture}
}
\caption{}  \label{fig_MGaa}
\end{subfigure}

\vspace{1cm}
\begin{subfigure}{0.40\textwidth}
\centering
\resizebox{\linewidth}{!}{
\begin{tikzpicture}[scale=1.0, shorten >=1pt,auto,node distance=2.8cm, semithick]
                    
\tikzset{vertex/.style = {inner sep=0.4pt}}
\tikzset{edge/.style = {->,> = latex'}}
 
\node[vertex] (1) at  (0,0) {$X_1^1$};
\node[vertex] (3) at  (1.5,0) {$X_3^1$};
\node[vertex] (2) at  (0.75,-1.5) {$X_2^1$};
\node[vertex] (7) at  (2.75,-1.5) {$T_1$};

\draw[edge] (1) to (2);

\node[vertex] (4) at  (4,0) {$X_3^2$};
\node[vertex] (6) at  (5.5,0) {$X_1^2$};
\node[vertex] (5) at  (4.75,-1.5) {$X_2^2$};

\draw[edge] (5) to (4);

\draw[edge] (7) to (2);
\draw[edge] (7) to (3);
\draw[edge] (7) to (5);
\draw[edge] (7) to (4);
\end{tikzpicture}
}
\caption{}  \label{fig_MGbb}
\end{subfigure}
\hspace{2cm}\begin{subfigure}{0.19\textwidth}
\centering
\resizebox{\linewidth}{!}{
\begin{tikzpicture}[scale=1.0, shorten >=1pt,auto,node distance=2.8cm, semithick]
                    
\tikzset{vertex/.style = {inner sep=0.4pt}}
\tikzset{edge/.style = {->,> = latex'}}
 
\node[vertex] (1) at  (0,0) {$X_1$};
\node[vertex] (3) at  (1.5,0) {$X_3$};
\node[vertex] (2) at  (0.75,-1.5) {$X_2$};

\draw[edge] (1) to (2);
\draw[edge] (2) to (3);

\node[vertex] (7) at  (2.15,-1.3) {$T_1$};
\draw[edge] (7) to (2);
\draw[edge] (7) to (3);

\end{tikzpicture}
}
\caption{}  \label{fig_MGcc}
\end{subfigure}

\caption{The mixture graph versus the fused graph. We plot the two DAGs over $\bm{Z}$ in (a) next to each to create $\mathbb{M}$ in (b). We have $\{X_1^1, X_1^2\} \ci_d \{X_3^1, X_3^2\}$ in $\mathbb{M}$ which implies $X_1 \ci X_3$. The fused graph $\mathbb{F}$ in (c) however does not imply the independence relation.}\label{fig_MGoo}
\end{figure*}
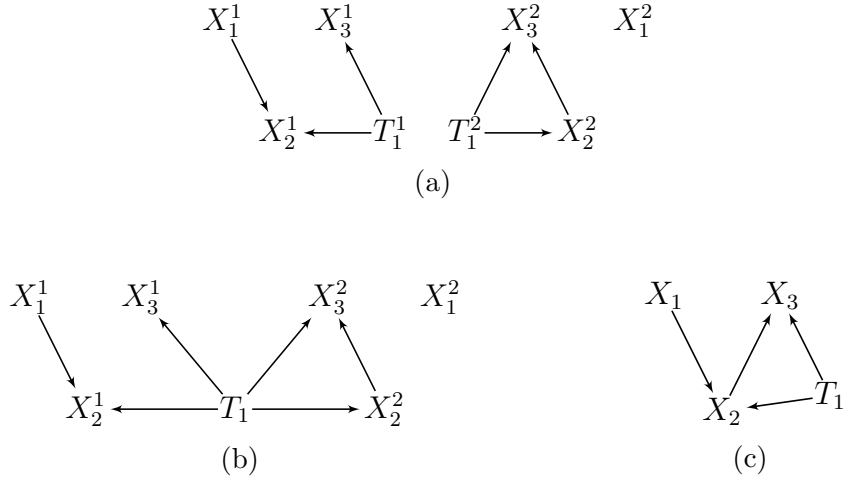

We adopt the following procedure:
\begin{enumerate}
\item Plot each of the $q$ DAGs in $\mathcal{G}$ adjacent to each other. 
\item Combine the vertices $T_i^\prime$ into a single vertex $T_i$ for each $T_i \in \bm{T}$.
\item Add additional directed edges so that the children of $T_i$ correspond to $\textnormal{Ch}^\prime(T_i)$ for each $T_i \in \bm{T}$.
\end{enumerate}
Denote the resultant graph as the \textit{mixture graph} $\mathbb{M}$. If $\bm{A} \subseteq \bm{T}$, then $\bm{A}^\prime = \bm{A}$ in $\mathbb{M}$ due to step 2 above. We can read off the implied CI relations from $\mathbb{M}$ by utilizing d-separation across groups of vertices rather than just singletons.
\begin{theorem1} \label{thm_DMP}
(Global Markov property) Let $\bm{A},\bm{B},\bm{C}$ denote disjoint subsets of $\bm{Z}$. If $\bm{A}^\prime \ci_d \bm{B}^\prime | \bm{C}^\prime$ in $\mathbb{M}$, then $\bm{A} \ci \bm{B} | \bm{C}$.
\end{theorem1}
\noindent We refer to the reverse direction as d-separation faithfulness with respect to $\mathbb{M}$. The result improves that of \citep{Spirtes94} (Appendix \ref{sec_spirtes}) and extends that of \citep{Saeed20} when $\bm{T}$ is partially observed, continuous or multivariate. We provide an example in Figure \ref{fig_MGoo}. Suppose $\bm{T}=\{T_1\}$ indexes two DAGs. We plot the two DAGs next to each other in Figure \ref{fig_MGaa} and combine the vertices associated with $\bm{T}$ as in Figure \ref{fig_MGbb}. We have $X_1^1 \rightarrow X_2^1$ in the first DAG and $X_2^2 \rightarrow X_3^2$ in the second; however, we do not have the directed path $X_1^j \rightarrow X_2^j \rightarrow X_3^j$ in either DAG. We also have the relation $\{X_1^1,X_1^2\} \ci_d \{X_3^1,X_3^2\}$, so $\mathbb{M}$ implies $X_1 \ci X_3 $ per Theorem \ref{thm_DMP}. 

\section{Causal Inference over Mixtures}\label{sec_algo}

\subsection{Fused Graph}
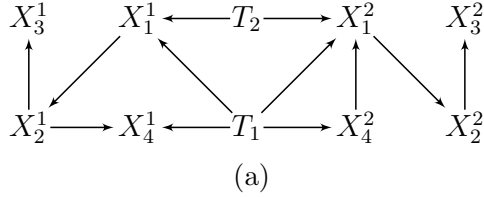
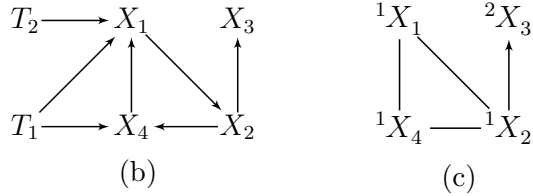
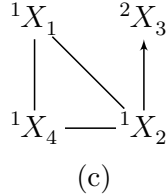
\begin{figure*}
\centering
\begin{subfigure}{0.42\textwidth}
\centering
\resizebox{\linewidth}{!}{
\begin{tikzpicture}[scale=1.0, shorten >=1pt,auto,node distance=2.8cm, semithick]
                    
\tikzset{vertex/.style = {inner sep=0.4pt}}
\tikzset{edge/.style = {->,> = latex'}}
 
\node[vertex] (1) at  (0.5,0) {$X_3^1$};
\node[vertex] (2) at  (2,-1.5) {$X_4^1$};
\node[vertex] (3) at  (2,0) {$X_1^1$};
\node[vertex] (4) at  (0.5,-1.5) {$X_2^1$};

\draw[edge] (3) to (4);
\draw[edge] (4) to (2);
\draw[edge] (4) to (1);

\node[vertex] (8) at  (3.5,-1.5) {$T_1$};
\draw[edge] (8) to (2);
\draw[edge] (8) to (3);

\node[vertex] (9) at  (3.5,0) {$T_2$};
\draw[edge] (9) to (3);

\node[vertex] (1) at  (5,0) {$X_1^2$};
\node[vertex] (2) at  (6.5,-1.5) {$X_2^2$};
\node[vertex] (3) at  (6.5,0) {$X_3^2$};
\node[vertex] (4) at  (5,-1.5) {$X_4^2$};

\draw[edge] (1) to (2);
\draw[edge] (2) to (3);
\draw[edge] (4) to (1);
\draw[edge] (8) to (1);
\draw[edge] (8) to (4);
\draw[edge] (9) to (1);
\end{tikzpicture}
}
\caption{}  \label{fig_mix_DAGs}
\end{subfigure}
\vspace{10mm}

\begin{subfigure}{0.225\textwidth}
\centering
\resizebox{\linewidth}{!}{
\begin{tikzpicture}[scale=1.0, shorten >=1pt,auto,node distance=2.8cm, semithick]
                    
\tikzset{vertex/.style = {inner sep=0.4pt}}
\tikzset{edge/.style = {->,> = latex'}}
 
\node[vertex] (1) at  (0,0) {$X_1$};
\node[vertex] (2) at  (1.5,-1.5) {$X_2$};
\node[vertex] (3) at  (1.5,0) {$X_3$};
\node[vertex] (4) at  (0,-1.5) {$X_4$};
\node[vertex] (8) at  (-1.5,-1.5) {$T_1$};
\node[vertex] (9) at  (-1.5,0) {$T_2$};

\draw[edge] (1) to (2);
\draw[edge] (2) to (3);
\draw[edge] (2) to (4);
\draw[edge] (4) to (1);
\draw[edge] (8) to (1);
\draw[edge] (8) to (4);
\draw[edge] (9) to (1);

\end{tikzpicture}
}
\caption{}  \label{fig_mix_mother}
\end{subfigure}
\hspace{10mm}
\begin{subfigure}{0.155\textwidth}
\centering
\resizebox{\linewidth}{!}{
\begin{tikzpicture}[scale=1.0, shorten >=1pt,auto,node distance=2.8cm, semithick]
                    
\tikzset{vertex/.style = {inner sep=0.4pt}}
\tikzset{edge/.style = {->,> = latex'}}
 
\node[vertex] (1) at  (0,0) {$\tensor[^1]{X}{_1}$};
\node[vertex] (2) at  (1.5,-1.5) {$\tensor[^1]{X}{_2}$};
\node[vertex] (3) at  (1.5,0) {$\tensor[^2]{X}{_3}$};
\node[vertex] (4) at  (0,-1.5) {$\tensor[^1]{X}{_4}$};

\draw[edge,-] (1) to (2);
\draw[edge] (2) to (3);
\draw[edge,-] (2) to (4);
\draw[edge,-] (4) to (1);

\end{tikzpicture}
}
\caption{}  \label{fig_mix_MG}
\end{subfigure}

\caption{We have the mixture graph in (a) and the fused graph in (b). Subfigure (c) contains $\mathbb{F}^*$ as well as wave information.}\label{fig_mix}
\end{figure*}

We construct a \textit{fused graph} $\mathbb{F}$ as follows. Create a vertex for every variable in $\bm{Z}$. Draw a directed edge $Z_i \rightarrow Z_j$ if and only if $Z_i^\prime$ is a direct cause of $Z_j^\prime$ in $\mathbb{M}$, so that $\mathbb{F}$ may contain cycles. The fused graph is more intuitive than the mixture graph because $\mathbb{F}$ summarizes cycles in one directed graph. We will utilize the global Markov property of $\mathbb{M}$ in order to recover a mixed graph $\mathbb{F}^*$ summarizing the ancestral relations in $\mathbb{F}$. For example, suppose we have the mixture graph drawn in Figure \ref{fig_mix_DAGs}. We consider a cycle involving $\{X_1,X_2,X_4\}$ and consider two slow causal relations: $X_2 \rightarrow X_4$ and $X_4 \rightarrow X_1$. We thus have $X_2 \rightarrow X_4$ in the first DAG in $\mathbb{M}$, but $X_4$ is overwritten by this causal relation, so we do not observe $X_4 \rightarrow X_1$. Likewise, we have $X_4 \rightarrow X_1$ in the second DAG, but $X_2$ is overwritten, so we do not observe $X_2 \rightarrow X_4$. We therefore cannot observe $X_2$ causing $X_1$ in either DAG due to the two rate limiting steps even though $X_2$ causes $X_1$ in the cycle involving $\{X_1,X_2,X_4\}$. Moreover, if we intervene on the value of $X_2$, then $X_2$ cannot be overwritten in the second DAG, so we would observe $X_2$ causing $X_1$. Now we have also drawn out $\mathbb{F}$ in Figure \ref{fig_mix_mother}. $X_2$ is an ancestor of $X_1$ in $\mathbb{F}$ even though $\{X_2^1,X_2^2\}$ is not an ancestor of $\{X_1^1,X_1^2\}$ in $\mathbb{M}$. Discovering $\mathbb{F}^*$ thus allows us to infer cycles that are not present within $\mathbb{M}$ but exist once the DAGs are combined in $\mathbb{F}$.

\subsection{Strategy}
We unfortunately cannot recover non-ancestral relations using a CI oracle alone (Appendix \ref{app_imp}), but we can recover ancestral relations. We therefore rely on additional time information to orient arrowheads by utilizing \textit{longitudinal data}, or data arising from a \textit{longitudinal density}. We can partition the observed variables into $w$ sets or \textit{waves} so that $\bm{O} = \cup_{k=1}^w\tensor[^k]{\bm{O}}{}$. We have the following definition:
\begin{definition1} (Longitudinal density) A longitudinal density is a density $f(\cup_{k=1}^w\tensor[^k]{\bm{O}}{}, \bm{L}, \bm{S})$ that factorizes according to Equation \eqref{eq_fac_endM} such that no variable in wave $j$ is an ancestor of a variable in wave $i<j$ and $w \geq 2$.
\end{definition1}
\noindent Causation proceeds forward in time, so no variable in wave $j$ can be an ancestor of a variable in wave $i<j$.

If $\bm{Y} \subseteq \bm{O}$, then let $\tensor[^a]{\bm{Y}}{}$ and $\tensor[^a]{\bm{Y}}{}^\prime$ denote $\bm{Y} \cap\tensor[^a]{\bm{O}}{}$ and $[\tensor[^a]{\bm{Y}}{}]^\prime$, respectively. We write $\tensor*[^{c}_d]{\textnormal{Adj}}{}_{\mathbb{F}^*}(\tensor[^a]{O}{_i})$ to mean those variables between waves $c$ and $d$ inclusive that are adjacent to $\tensor[^a]{O}{_i}$ in $\mathbb{F}^*$. We will specifically construct $\mathbb{F}^*$ with the following adjacencies:
\begin{listn} (Adjacency Interpretations) \\ \label{list_adj}
\begin{enumerate}
\item If we have $\tensor[^a]{O}{_i} * \!\! - \!\! * \tensor[^b]{O}{_j}$ (with possibly $a=b$), then $\tensor[^a]{O}{}_i^\prime \not \ci_d \tensor[^b]{O}{}_j^\prime | \bm{W}^\prime \cup \bm{S}^\prime$ in $\mathbb{M}$ for all $\bm{W} \subseteq \tensor*[^{a}_b]{\textnormal{Adj}}{_{\mathbb{F}^*}}(\tensor[^a]{O}{_i}) \setminus \tensor[^b]{O}{_j}$ and all $\bm{W} \subseteq \tensor*[^{a}_b]{\textnormal{Adj}}{_{\mathbb{F}^*}}(\tensor[^b]{O}{_j}) \setminus \tensor[^a]{O}{_i}$.
\item If we do not have  $\tensor[^a]{O}{_i} * \!\! - \!\! * \tensor[^b]{O}{_j}$ (with possibly $a=b$), then $\tensor[^a]{O}{}_i^\prime \ci_d \tensor[^b]{O}{}_j^\prime | \bm{W}^\prime \cup \bm{S}^\prime$ in $\mathbb{M}$ for some $\bm{W} \subseteq \bm{O} \setminus \{ \tensor[^a]{O}{_i}, \tensor[^b]{O}{_j}\}$ between waves $a$ and $b$ inclusive.
\end{enumerate}
\end{listn}
\noindent The endpoints of $\mathbb{F}^*$ have the following modified interpretations:
\begin{listn} (Endpoint Interpretations) \\ \label{list_eps2}
\begin{enumerate}
\item If $O_i * \!\! \rightarrow O_j$, then $O_j \not \in \textnormal{Anc}_{\mathbb{F}}(O_i)$.
\item If $O_i * \!\! - O_j$, then $O_j \in \textnormal{Anc}_{\mathbb{F}}(O_i \cup \bm{S})$.
\end{enumerate}
\end{listn}
\noindent The arrowheads do not take into account selection variables because we often cannot a priori specify whether a variable is an ancestor of $\bm{S}$ in $\mathbb{F}$ using either wave information or other prior knowledge in practice. We draw an example of $\mathbb{M}$ in Figure \ref{fig_mix_DAGs}, its fused graph $\mathbb{F}$ in Figure \ref{fig_mix_mother} and the corresponding mixed graph $\mathbb{F}^*$ in Figure \ref{fig_mix_MG}, where $\bm{O} =\bm{X}$, $\bm{L}=\{T_1,T_2\}$, $\bm{S}=\emptyset$ and $w=2$.

\subsection{Algorithm} \label{sec_alg}

We cannot apply an existing constraint-based algorithm like FCI on data arising from a mixture of DAGs and expect to recover a partially oriented $\mathbb{F}^*$ (Appendix \ref{app_fail}). We therefore propose a new algorithm called Causal Inference over Mixtures (CIM) which correctly recovers causal relations. We summarize the procedure in Algorithm \ref{alg_CIM}. 

\begin{algorithm}[]
 \KwData{CI oracle, waves $\mathcal{W}$, other prior information $\mathcal{P}$}
 \KwResult{partially oriented mixed graph $\widehat{\mathbb{F}}^*$}
 \BlankLine
 
Run Algorithm \ref{pc_skel}, a variant of PC-stable's skeleton discovery procedure.\\ \label{alg_skeleton}

If we have $O_i * \!\! \linecirc O_j$ and $O_i$ lies within an earlier wave than $O_j$ according to $\mathcal{W}$ or $O_j$ cannot be an ancestor of $O_i$ according to $\mathcal{P}$, then orient $O_i * \!\! \linecirc O_j$ as $O_i * \!\! \rightarrow O_j$ in $\widehat{\mathbb{F}}^*$. \label{alg_pk}

If we have $\tensor[^a]{O}{}_i * \!\!\rightarrow O_j * \!\! - \!\! * \tensor[^b]{O}{}_k$ with $\tensor[^a]{O}{}_i$ and $\tensor[^b]{O}{}_k$ non-adjacent, $O_j \not \in \textnormal{Sep}(\tensor[^a]{O}{}_i,\tensor[^b]{O}{}_k)$ and there exists another minimal separating set $\bm{W} \subseteq  \tensor*[^{a}_b]{\textnormal{Adj}}{_{\mathbb{F}^*}}(\tensor[^a]{O}{}_i)\setminus \tensor[^b]{O}{}_k$ or $\bm{W} \subseteq  \tensor*[^{a}_b]{\textnormal{Adj}}{_{\mathbb{F}^*}}(\tensor[^b]{O}{}_k)\setminus \tensor[^a]{O}{}_i$ containing $O_j$, then record $\bm{W}$ into $\textnormal{Sep2}(\tensor[^a]{O}{}_i,O_j, \tensor[^b]{O}{}_k)$. \label{alg_dsep}

If we have $O_i * \!\! \rightarrow O_j \circline \!\! * O_k$ with $O_i$ and $O_k$ non-adjacent, and either $O_j \in \textnormal{Sep}(O_i, O_k)$ or $\textnormal{Sep2}(O_i, O_j,O_k)$ is non-empty, then orient $O_j \circline \!\! * O_k$ as $O_j - \!\! * O_k$ in $\widehat{\mathbb{F}}^*$. \label{alg_tail1}

Execute the following orientation rule until no more edges can be oriented: if we have the sequence of vertices $\langle O_1, \dots, O_n \rangle$ such that $O_i - \!\! * O_{i+1}$ with $ 1 \leq i \leq n-1$, and we have $O_1 \circline \!\! * O_n$, then orient  $O_1 \circline \!\! * O_n$ as $O_1 - \!\! * O_n$ in $\widehat{\mathbb{F}}^*$. \label{alg_tail2}
 \BlankLine

 \caption{Causal Inference over Mixtures (CIM)} \label{alg_CIM}
\end{algorithm}

The CIM algorithm works as follows. First, CIM runs a variant of PC-stable's skeleton discovery procedure in order to discover adjacencies as well as minimal separating sets in Step \ref{alg_skeleton} \citep{Colombo14}. This step recovers the adjacencies with interpretations listed in List \ref{list_adj}. The algorithm stores the minimal separating sets in the array Sep so that $\textnormal{Sep}(O_i, O_k)$ contains a minimal separating set of $O_i$ and $O_k$, if such a set exists. CIM next adds arrowheads in Step \ref{alg_pk} using wave information from a longitudinal dataset with the list $\mathcal{W}$. If we have $\tensor[^a]{O}{_i} \circlinecirc \tensor[^b]{O}{_j}$ with $b > a$, then CIM orients $\tensor[^a]{O}{_i} \circarrow \tensor[^b]{O}{_j}$ because $\tensor[^b]{O}{_j} \not \in \textnormal{Anc}_{\mathbb{F}}(\tensor[^a]{O}{_i})$. We can orient additional arrowheads using other prior knowledge $\mathcal{P}$. Step \ref{alg_pk} orients many arrowheads in practice, so long as we have at least two waves of data and repeated measurements.

For every triple $O_i * \!\! \rightarrow O_j * \!\! - \!\! * O_k$ with $O_i$ and $O_k$ non-adjacent, CIM then attempts to find a minimal separating set that contains $O_j$ in Step \ref{alg_dsep}. These sets are important due to the following lemma which allows us to infer tails in Step \ref{alg_tail1}: 
\begin{lemma1}\label{lem_anc1}
Suppose $O_i^\prime \ci_d O_j^\prime|\bm{W}^\prime \cup \bm{S}^\prime$ in $\mathbb{M}$ but $O_i^\prime \not \ci_d O_j^\prime|\bm{V}^\prime \cup \bm{S}^\prime$ for every $\bm{V} \subset \bm{W}$. If $O_k \in \bm{W}$, then $O_k \in \textnormal{Anc}_{\mathbb{F}}(\{O_i, O_j\} \cup \bm{S})$.
\end{lemma1}
\noindent CIM finally adds some additional tails in Step \ref{alg_tail2} due to transitivity of the tails. The algorithm has the same polynomial time complexity as PC-stable due to Step \ref{alg_skeleton}.

We now formally claim that Algorithm \ref{alg_CIM} is sound:
\begin{theorem1} \label{thm_sound}
Suppose the longitudinal density $f(\cup_{k=1}^w\tensor[^k]{\bm{O}}{}, \bm{L}, \bm{S})$ factorizes according to Equation \eqref{eq_fac_endM}. Assume that all arrowheads deduced from $\mathcal{P}$ are correct. Then, under d-separation faithfulness w.r.t. $\mathbb{M}$, the CIM algorithm returns the mixed graph $\mathbb{F}^*$ partially oriented.
\end{theorem1}

\section{Experiments} \label{sec_exp}

We had two overarching goals: (1) evaluate the performance of CIM against other constraint-based algorithms using real data, and (2) determine if we can reconstruct the real data results using synthetic data sampled from a mixture of DAGs. We therefore utilized the setup described below.

\subsection{Algorithms}
We compared the following five constraint-based algorithms in recovering the ancestral and nonancestral relations in $\mathbb{F}$: CIM, PC, FCI, RFCI and CCI. We equipped all algorithms with a nonparametric CI test called GCM \citep{Shah20} and fixed $\alpha=0.01$ across all experiments. We gave all algorithms the same wave information during skeleton discovery in order to orient arrowheads between the waves. The algorithms perform much worse without the additional knowledge.

\subsubsection{Metrics}
Let tails refer to positives and arrowheads to negatives. CIM only infers tails, so we cannot compute the number of true and false negatives. We can however compute the number of true positives and false positives. 

We therefore evaluated the algorithms using \textit{sensitivity} and \textit{fallout}. The sensitivity is defined as $TP/P$, where $TP$ refers to true positives and $P$ to positives. The fallout is defined as $FP/N$, where $FP$ refers to false positives and $N$ to negatives. A tail in place of an arrowhead corresponds to a false positive. 

The receiver operating characteristic (ROC) curve plots sensitivity against the fallout. Perfect accuracy corresponds to a sensitivity of one and a fallout of zero at the upper left hand corner of the ROC curve. Constraint-based algorithms do not output a continuous score required to compute the area under the ROC curve, but we can assess \textit{overall performance} using the Euclidean distance from the upper left hand corner \citep{Perkins06}. 

\subsection{Real Data}

\subsubsection{Framingham Heart Study} \label{sec_real}

We first evaluated the algorithms on real data. We considered the Framingham Heart Study, where investigators measured cardiovascular changes across time in residents of Framingham, Massachusetts \citep{Mahmood14}. The dataset contains three waves of data with 8 variables in each wave. We obtained 2019 samples after removing patients with missing values. 

The dataset contains the following known direct causal relations: (1) number of cigarettes per day causes heart rate via cardiac nicotonic acetylcholine receptors \citep{Aronow71,Levy71,Haass97}; (2) age causes systolic blood pressure due to increased large artery stiffness \citep{Pinto07,Safar05}; (3) age causes cholesterol levels due to changes in cholesterol and lipoprotein metabolism \citep{Parini99}; (4) BMI causes number of cigarettes per day because smoking cigarettes is a common weight loss strategy \citep{Jo02,Chiolero08}; (5) systolic blood pressure causes diastolic blood pressure and vice versa by definition, because both quantities refer to pressure in the same arteries at different points in time. We can compute sensitivity using this information.

\begin{figure*}
\centering
\begin{subfigure}{0.32\textwidth}
  \centering
  \includegraphics[width=1\linewidth]{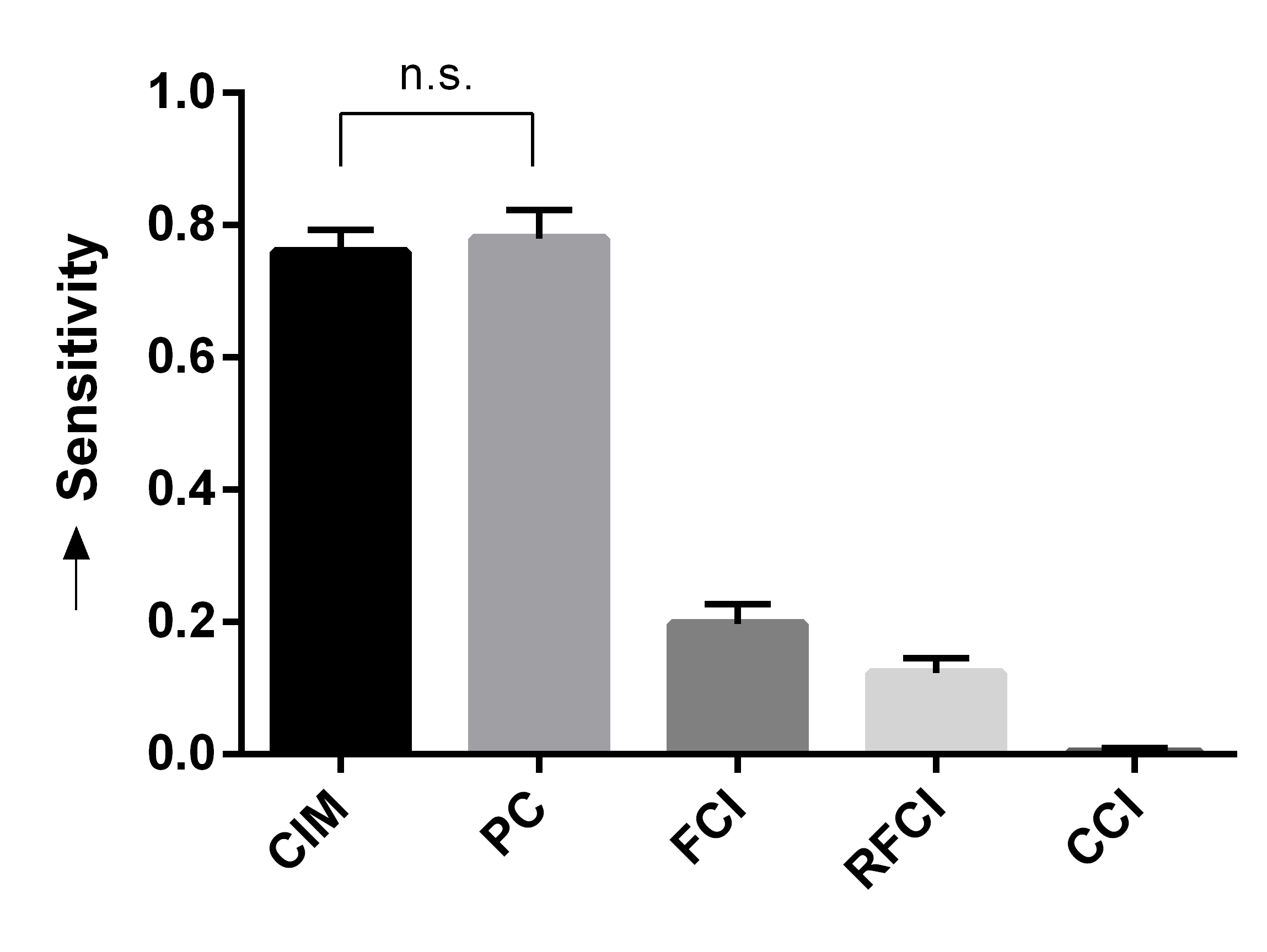}
  \caption{}
  \label{fig_FHS:sens}
\end{subfigure}
\begin{subfigure}{0.32\textwidth}
  \centering
  \includegraphics[width=1\linewidth]{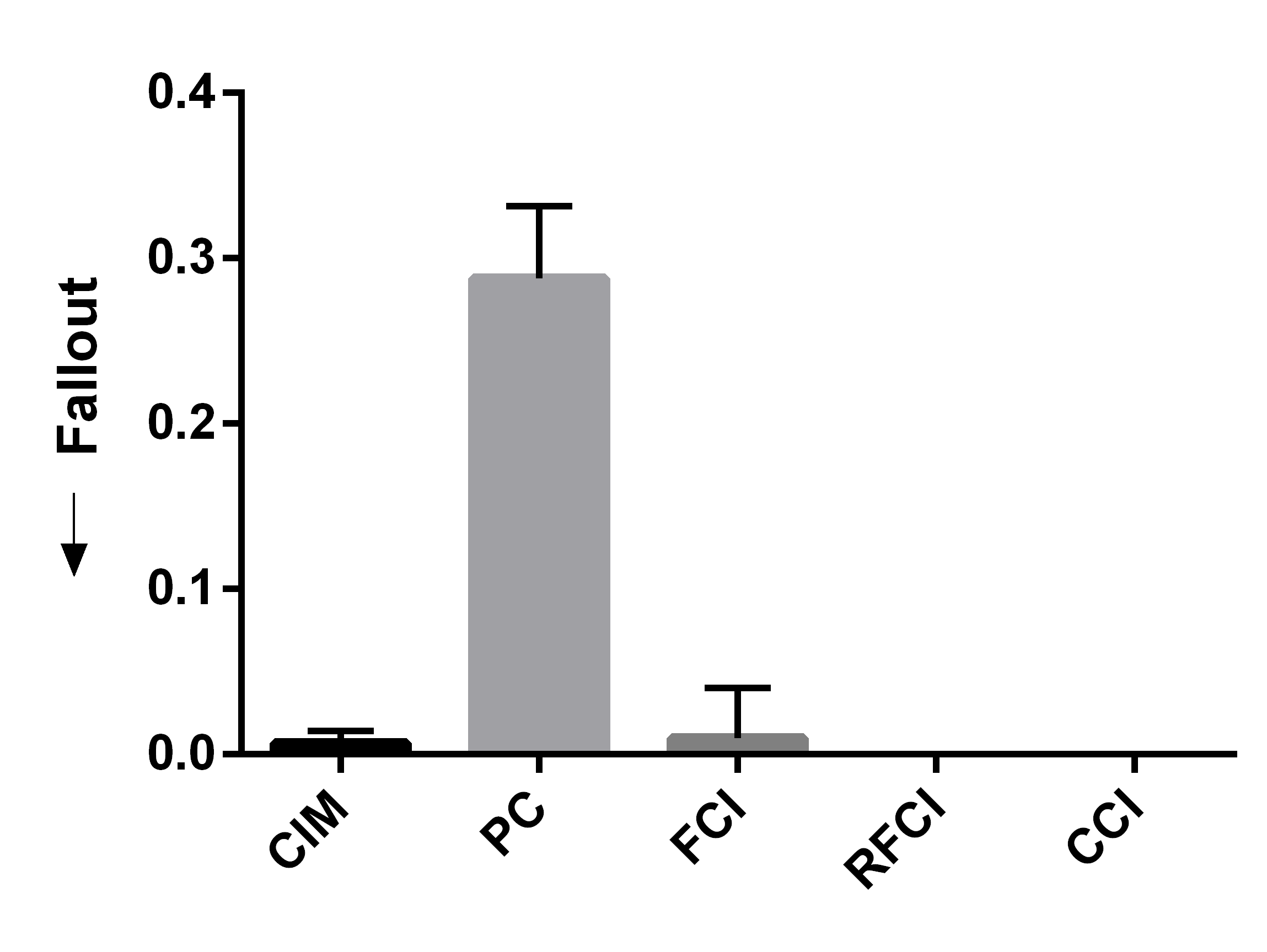}
  \caption{}
  \label{fig_FHS:fo}
\end{subfigure}
\begin{subfigure}{0.32\textwidth}
  \centering
  \includegraphics[width=1\linewidth]{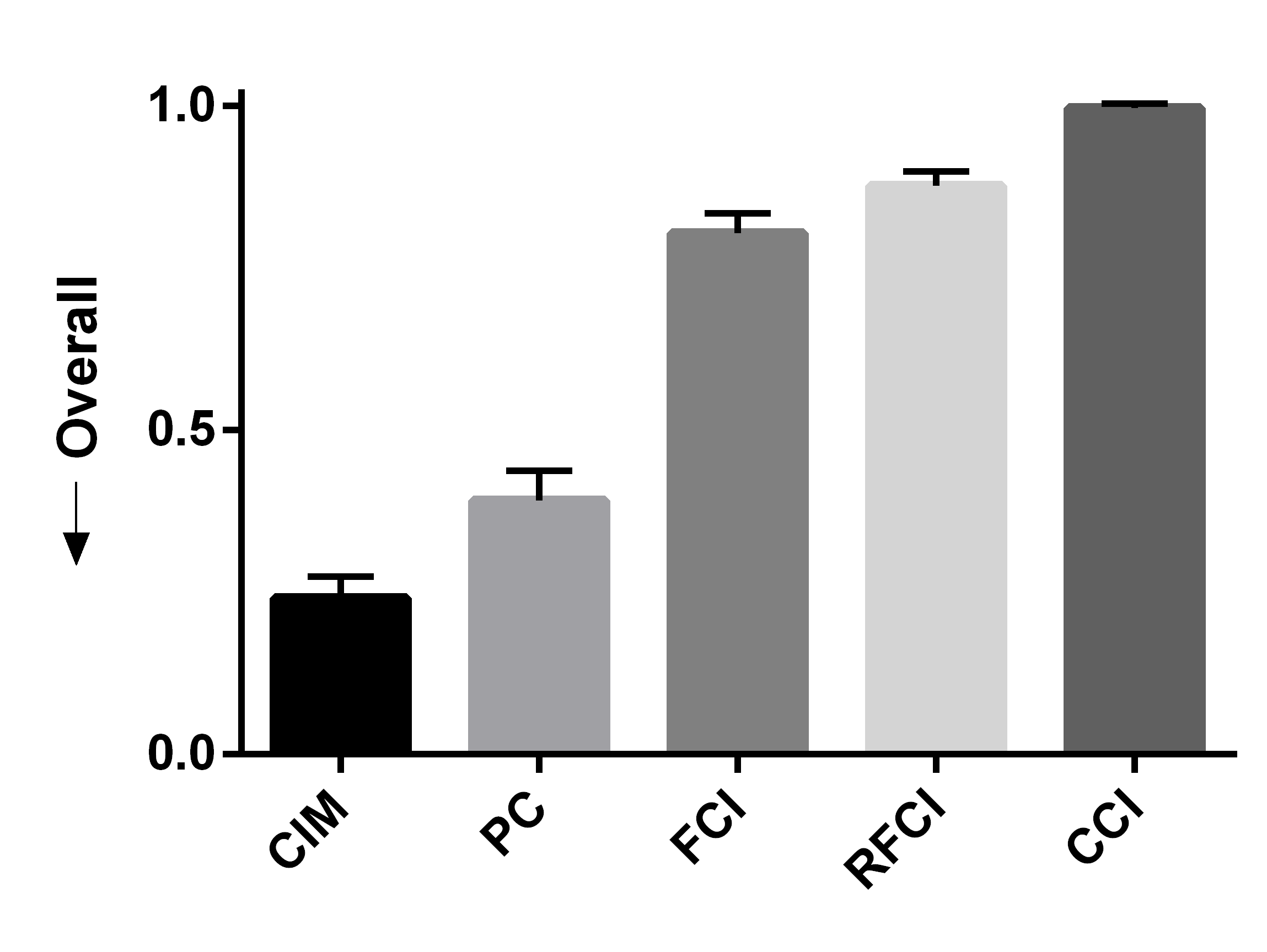}
  \caption{}
  \label{fig_FHS:oa}
\end{subfigure}

\begin{subfigure}{0.32\textwidth}
  \centering
  \includegraphics[width=1\linewidth]{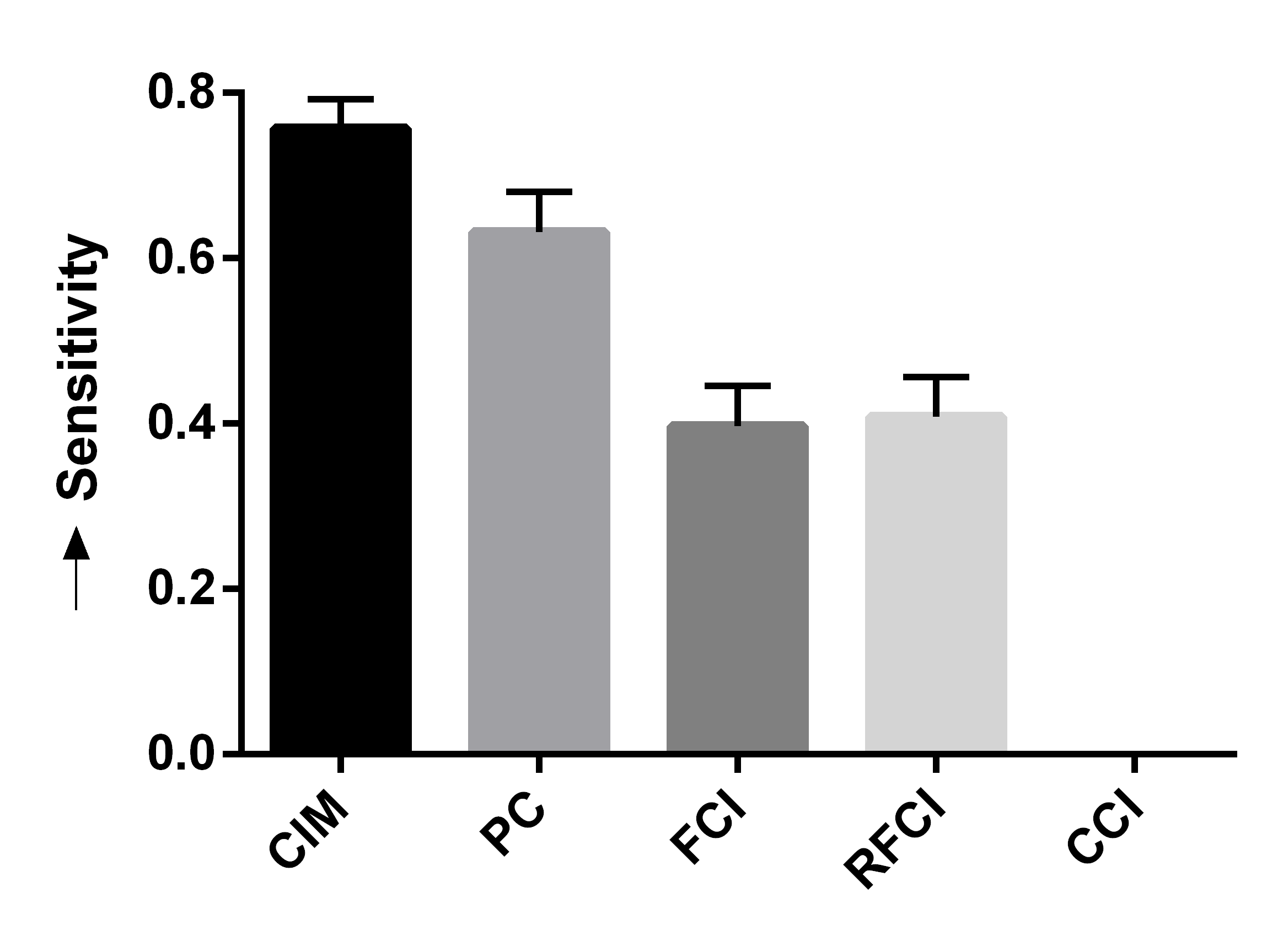}
  \caption{}
  \label{fig_STARD:sens}
\end{subfigure}
\begin{subfigure}{0.32\textwidth}
  \centering
  \includegraphics[width=1\linewidth]{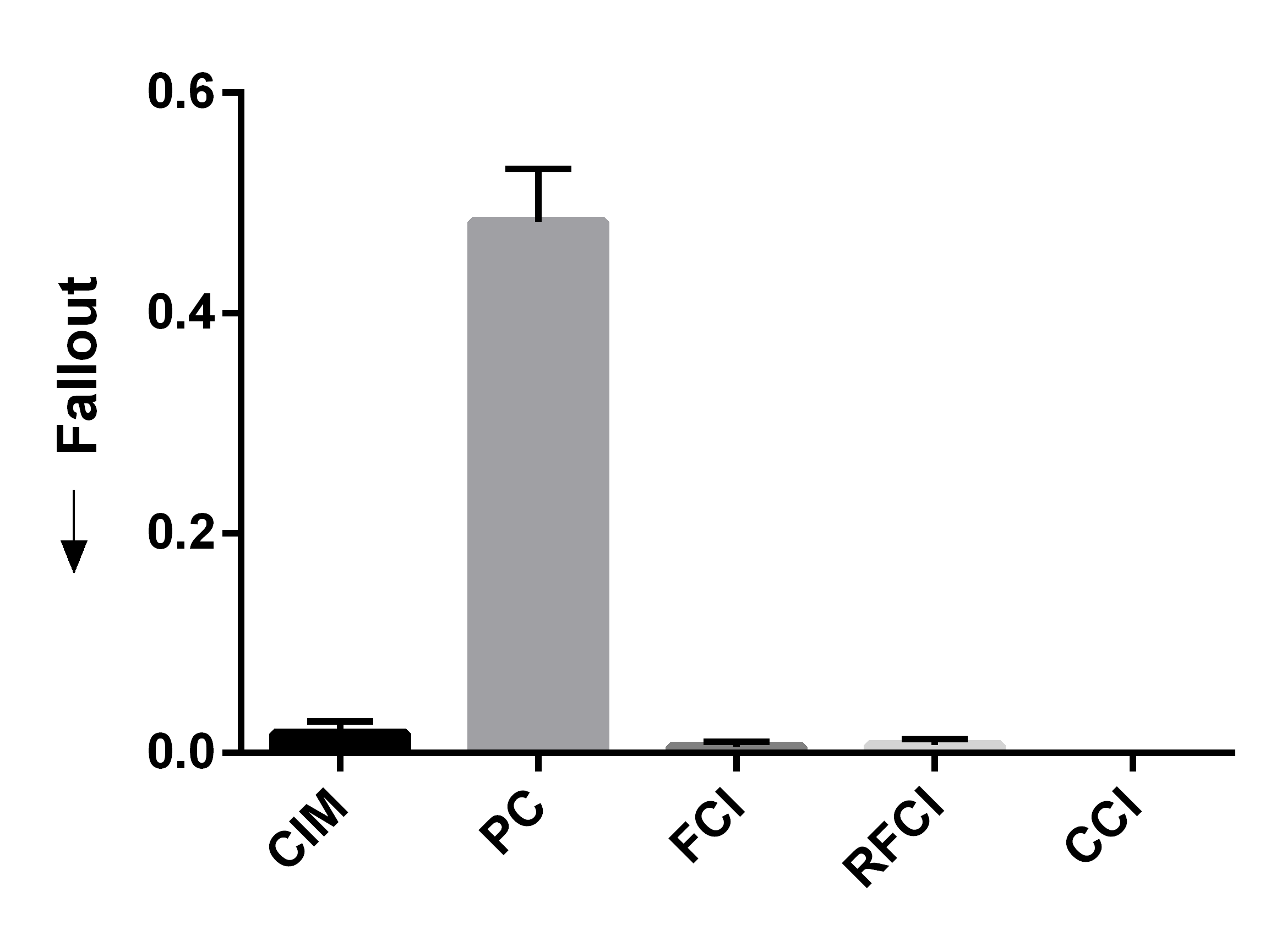}
  \caption{}
  \label{fig_STARD:fo}
\end{subfigure}
\begin{subfigure}{0.32\textwidth}
  \centering
  \includegraphics[width=1\linewidth]{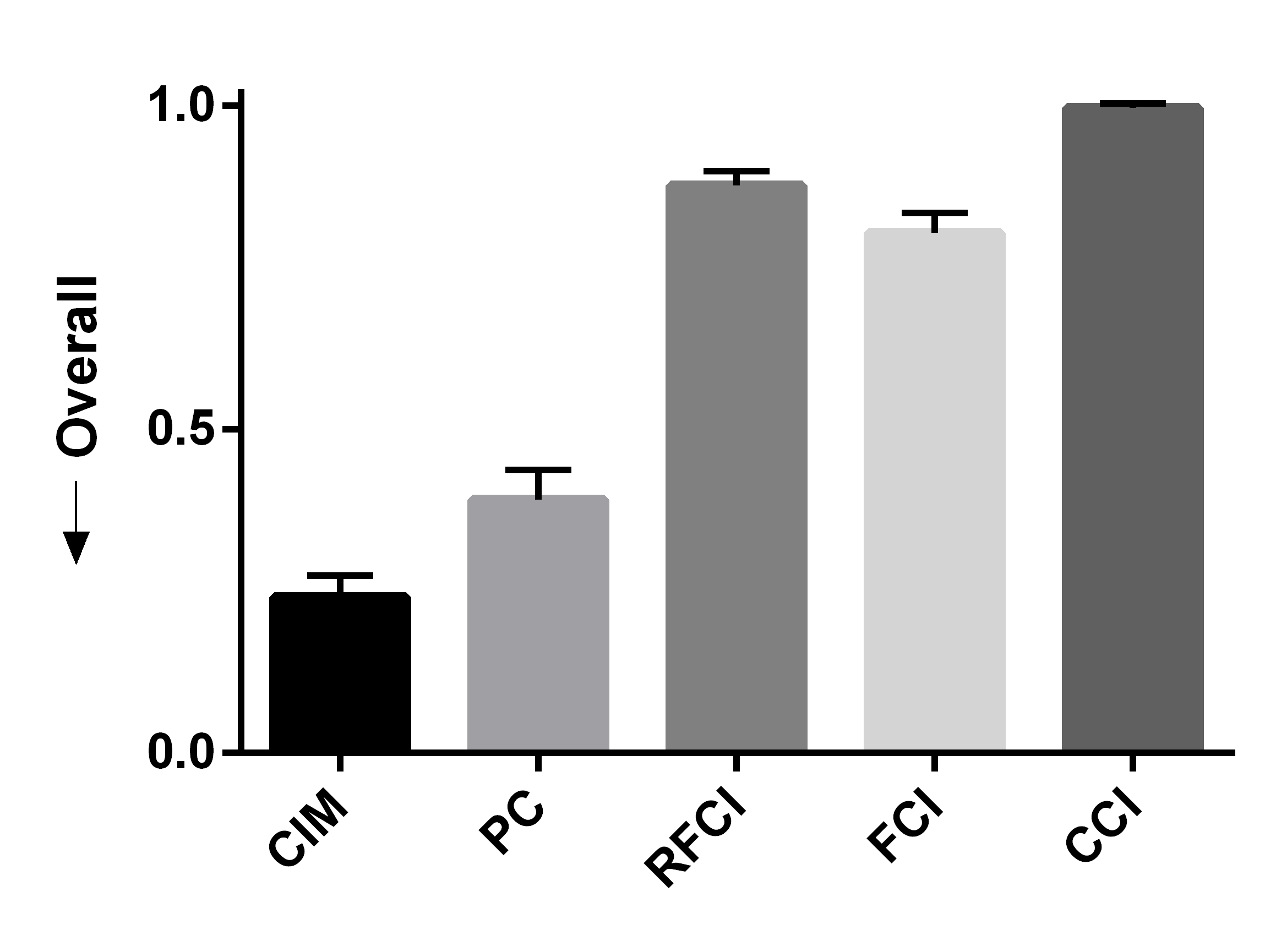}
  \caption{}
  \label{fig_STARD:oa}
\end{subfigure}

\begin{subfigure}{0.32\textwidth}
  \centering
  \includegraphics[width=1\linewidth]{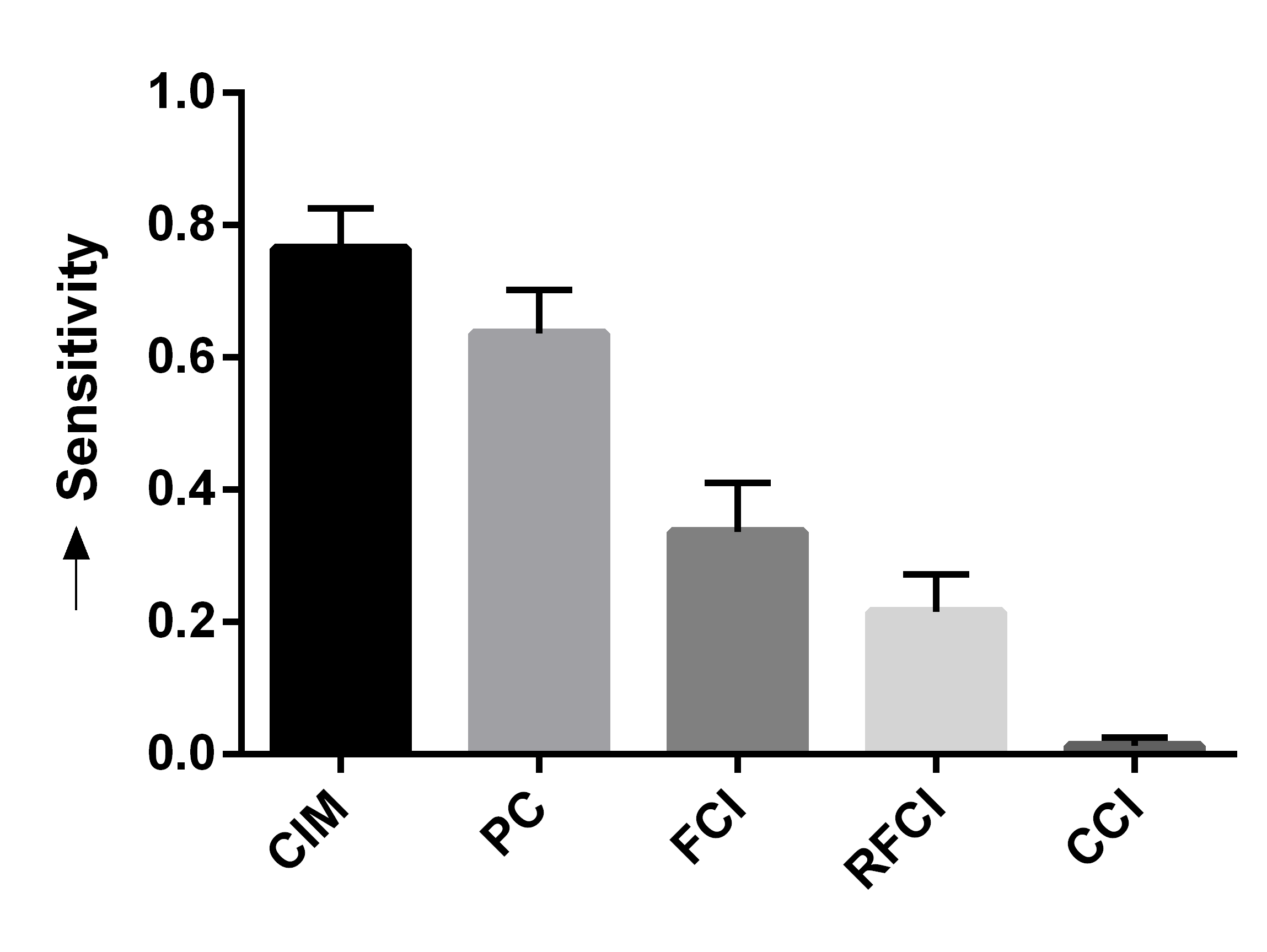}
  \caption{}
  \label{fig_synth:sens}
\end{subfigure}
\begin{subfigure}{0.32\textwidth}
  \centering
  \includegraphics[width=1\linewidth]{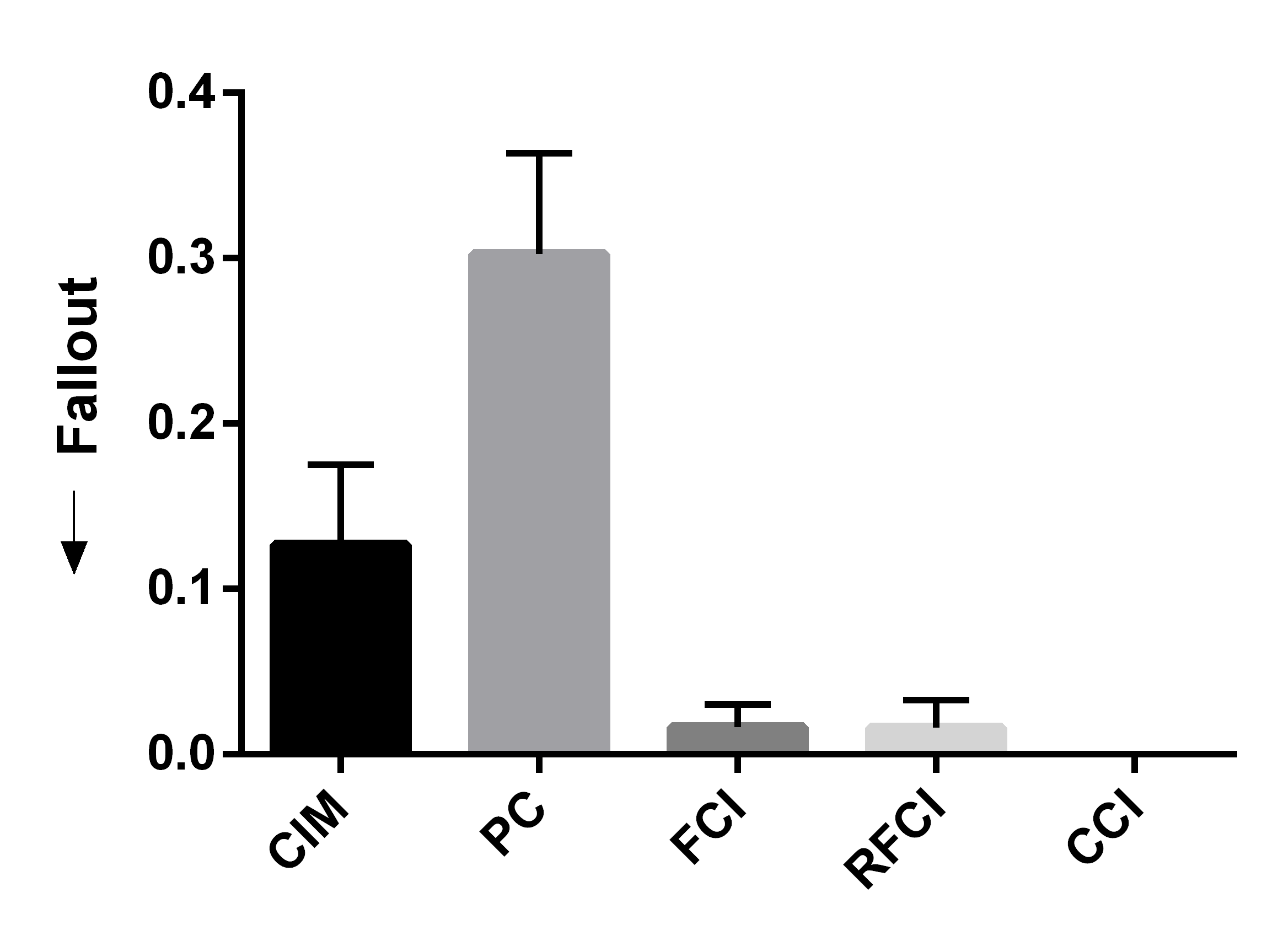}
  \caption{}
  \label{fig_synth:fo}
\end{subfigure}
\begin{subfigure}{0.32\textwidth}
  \centering
  \includegraphics[width=1\linewidth]{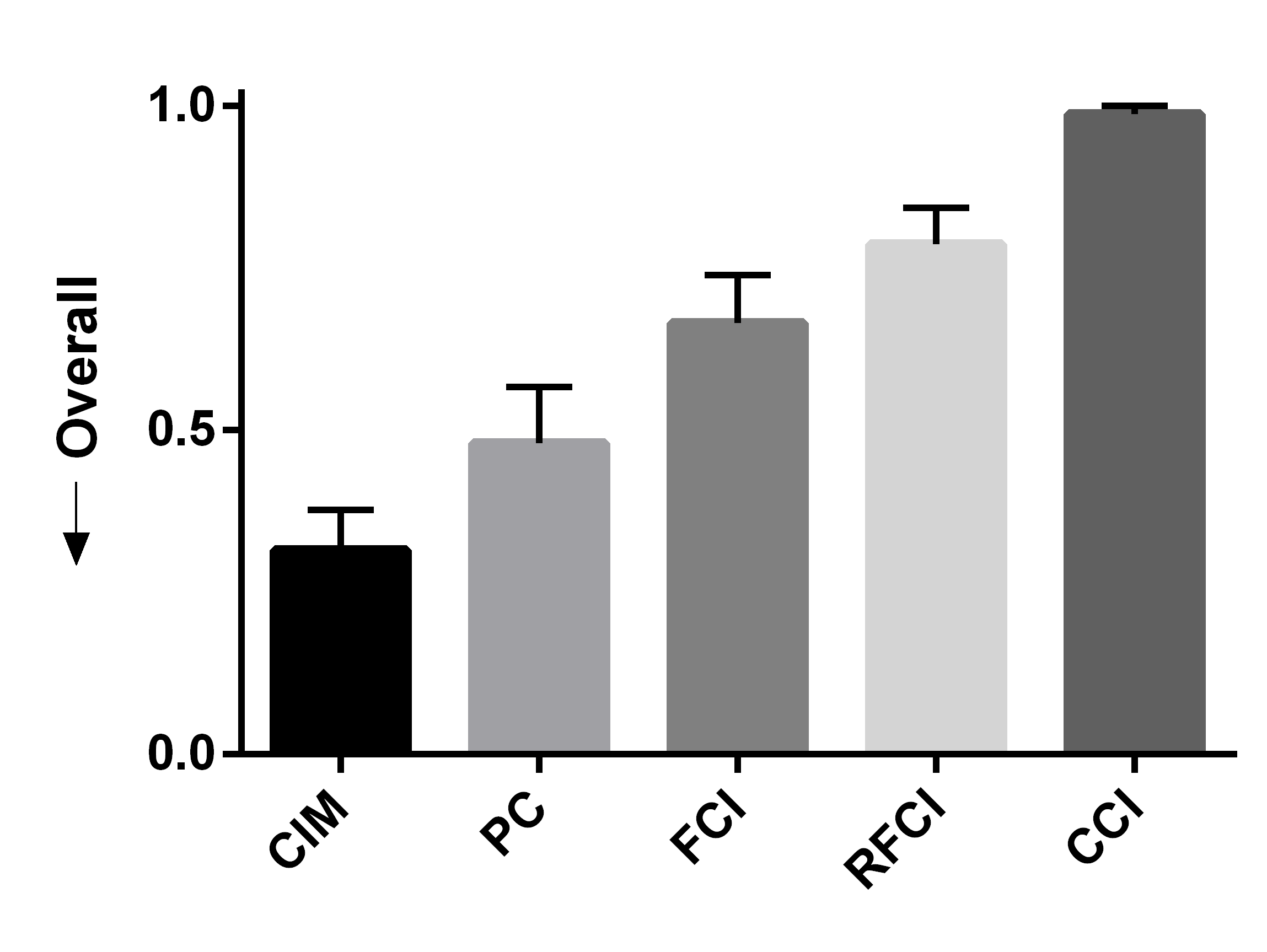}
  \caption{}
  \label{fig_synth:oa}
\end{subfigure}
\caption{Results for FHS in (a, b, c), STAR$^*$D in (d, e, f) and the synthetic data in (g, h, i). Bar heights represent empirical means and error bars their 95\% confidence intervals. An up-pointing arrow means higher is better and a down-pointing arrow means lower is better. CIM achieves higher sensitivity in (a, d, g) while maintaining a low fallout in (b, e, h). CIM performs the best overall in all cases as shown in (c, f, i).} \label{fig_FHS}
\end{figure*}

We summarize the results over 50 bootstrapped datasets in Figures \ref{fig_FHS} (a, b, c). We first evaluated sensitivity by running the algorithms using the full wave information. RFCI, FCI and CCI oriented few tails overall, so they obtained lower sensitivity scores (Figure \ref{fig_FHS:sens}). PC and CIM had similar sensitivities (t=-0.80, p=0.43). We next combined waves 2 and 3, so that the algorithms could incorrectly orient tails backwards in time. CIM made fewer errors than PC as indicated by a lower fallout (Figure \ref{fig_FHS:fo}, t=-11.85, p=5.37E-16). FCI, RFCI and CCI also achieved low fallout scores, but they again did not orient many tails to begin with. CIM therefore obtained the best overall score when we combined sensitivity and fallout (Figure \ref{fig_FHS:oa}, t=-5.60, p=9.70E-7). We conclude that both CIM and PC orient many tails, but CIM makes fewer errors as evidenced by its high sensitivity and low fallout. We therefore prefer CIM in this dataset.

\subsubsection{Sequenced Treatment Alternatives to Relieve Depression Trial}

We next analyzed Level 1 of the Sequenced Treatment Alternatives to Relieve Depression (STAR$^*$D) trial \citep{Sinyor10}. Investigators gave patients an antidepressant called citalopram and then tracked their depressive symptoms using a standardized questionnaire called QIDS-SR-16. We analyzed the 9 QIDS-SR-16 sub-scores measuring components of depression at weeks 0, 2 and 4. We also included age and gender in the first wave. The dataset contains 2043 subjects after removing subjects with missing values.

The 9 QIDS-SR-16 subscores include sleep, mood, appetite, concentration, self-esteem, thoughts of death, interest, energy and psychomotor changes. We asked a psychiatrist to identify direct ground truth causal relations among the subscores before we ran the experiments. The ground truth includes: (1) sleep causes mood \citep{Motomura17}; (2) energy causes psychomotor changes; (3) appetite causes energy; (4, 5) mood causes appetite and self-esteem \citep{Hepworth10}; (6) psychomotor changes cause concentration; (7, 8) mood and self-esteem cause thoughts of death \citep{Bhar08}. 

We summarize the sensitivity, fallout and overall performance over 50 bootstrapped datasets in Figures \ref{fig_FHS} (d, e, f). CIM achieved higher sensitivity than all other algorithms (Figure \ref{fig_STARD:sens}, t=5.66, p=7.86E-7). CIM also had a smaller fallout score compared to PC (Figure \ref{fig_STARD:fo}, t=-19.19, p$<$2.20E-16). CIM therefore obtained the highest overall score compared to the other algorithms (Figure \ref{fig_STARD:oa}, t=-14.95, p$<$2.20E-16). These results corroborate the superiority of CIM in a second real dataset.

\subsection{Synthetic Data}

We next sampled from a mixture of DAGs to see if we could replicate the real data results. We specifically instantiated a linear DAG with an expected neighborhood size of 2, $p=24$ vertices and linear coefficients drawn from Uniform($[-1,-0.25]\cup[0.25,1]$). We then uniformly instantiated $q=5$ to 15 binary variables for $\bm{T}$ and block randomized the edges in the DAG to each element of $\bm{T}$. We assigned the first 8 variables to wave 1, the second 8 to wave 2, and the third 8 to wave 3. We added a directed edge from the $n^\textnormal{th}$ variable in wave 1 to the $n^\textnormal{th}$ variable in wave 2, and similarly added the directed edges from wave 2 to wave 3 in order to model self-loops. We randomly selected a set of 0-2 latent common causes without replacement from $\bm{X}$, which we placed in $\bm{L}$ in addition to the variables in $\bm{T}$. We then selected a set of 0-2 selection variables $\bm{S}$ without replacement from the set $\bm{X} \setminus \bm{L}$. 

We uniformly instantiating the mixing probabilities $f(T_i = 0)$ and $f(T_i = 1)$ for each $T_i \in \bm{T}$. We then generated 2000 samples as follows. For each sample, we drew an instantiation $\bm{T}=\bm{t}$ according to $\prod_{i=1}^q f(T_i)$ and created a graph containing the union of the edges associated with those elements in $\bm{t}$ equal to one. We then sampled the resultant DAG using a multivariate Gaussian distribution. We finally removed the latent variables and introduced selection bias by removing the bottom $k^\textnormal{th}$ percentile for each selection variable, with $k$ chosen uniformly between 10 and 50.

We report the results in Figures \ref{fig_FHS} (g, h, i) after repeating the above process 50 times. We computed the sensitivity and fallout using the ground truth in waves 2 and 3. CIM achieved the highest sensitivity (Figure \ref{fig_synth:sens}, t=3.71, p=5.35E-4). PC obtained the second highest sensitivity, but CIM had a lower fallout than PC (Figure \ref{fig_synth:fo}, t=-4.63,p=2.72E-5). CIM ultimately achieved the best overall score (Figure \ref{fig_synth:oa}, t=-3.78,p=4.37E-4). We conclude that the synthetic data results mimic those seen with the real data.

\section{Conclusion} \label{sec_conc}

We proposed to model causal processes in biomedicine using a mixture of DAGs to accommodate non-stationary distributions and cycles. We then introduced a constraint-based algorithm called CIM to infer causal relations from data even with latent variables and selection bias. The CIM algorithm outperforms existing constraint-based algorithms across multiple metrics and datasets. CIM thus advances the state of the art in causal discovery from biomedical data.

\bibliographystyle{elsarticle-num-names}
\bibliography{biblio}

\section{Appendix}

\subsection{Proofs} \label{app_proofs}

\begin{reptheorem}{thm_DMP}
Let $\bm{A},\bm{B},\bm{C}$ denote disjoint subsets of $\bm{Z}$. If $\bm{A}^\prime \ci_d \bm{B}^\prime | \bm{C}^\prime$ in $\mathbb{M}$, then $\bm{A} \ci \bm{B} | \bm{C}$.
\end{reptheorem}
\begin{proof}
We first consider the moral graph of $\mathbb{M}$. Let $\widebar{\mathbb{M}}$ denote the moral graph of $\widebar{\textnormal{Anc}}_{\mathbb{M}}(\bm{A}^\prime\cup\bm{B}^\prime \cup \bm{C}^\prime )$, the smallest ancestral set of $\bm{A}^\prime\cup\bm{B}^\prime \cup \bm{C}^\prime$ in $\mathbb{M}$ such that, if $Z_i \in \widebar{\textnormal{Anc}}_{\mathbb{M}}(\bm{A}^\prime\cup\bm{B}^\prime \cup \bm{C}^\prime )$, then $Z_i^\prime \subseteq \widebar{\textnormal{Anc}}_{\mathbb{M}}(\bm{A}^\prime\cup\bm{B}^\prime \cup \bm{C}^\prime )$. We then consider a partition of the vertices $\ddot{\bm{A}} \cup \ddot{\bm{B}} \cup \bm{C}^\prime = \widebar{\textnormal{Anc}}_{\mathbb{M}}(\bm{A}^\prime\cup\bm{B}^\prime \cup \bm{C}^\prime )$ so that $\bm{A}^\prime \subseteq \ddot{\bm{A}}$, $\bm{B}^\prime \subseteq \ddot{\bm{B}}$, and $\ddot{\bm{A}}$, $\ddot{\bm{B}}$ and $\bm{C}^\prime$ are disjoint sets of vertices. We require that $\ddot{\bm{A}}$ and $\ddot{\bm{B}}$ be separated by $\bm{C}^\prime$ in $\widebar{\mathbb{M}}$; in other words, there does not exist an undirected path between $\ddot{\bm{A}}$ and $\ddot{\bm{B}}$ that is active given $\bm{C}^\prime$.

We now construct such a partition $(\ddot{\bm{A}},\ddot{\bm{B}})$. First set $\ddot{\bm{A}}$ to $\bm{A}^\prime$ and $\ddot{\bm{B}}$ to $\bm{B}^\prime$. We have $\bm{A}^\prime \ci_{d} \bm{B}^\prime | \bm{C}^\prime$ in $\mathbb{M}$ if and only if $\bm{A}^\prime$ and  $\bm{B}^\prime$ are separated by $\bm{C}^\prime$ in $\widebar{\mathbb{M}}$ (Lemma 2 in \citep{Lauritzen90}). $\ddot{\bm{A}}$ and  $\ddot{\bm{B}}$ are therefore separated by $\bm{C}^\prime$ in $\widebar{\mathbb{M}}$ at the moment. Now consider the set of vertices $\bm{H}^\prime = \widebar{\textnormal{Anc}}_{\mathbb{M}}(\bm{A}^\prime\cup\bm{B}^\prime \cup \bm{C}^\prime )\setminus (\bm{A}^\prime \cup \bm{B}^\prime \cup \bm{C}^\prime).$ We will put subsets of $\bm{H}^\prime$ into either $\ddot{\bm{A}}$ or $\ddot{\bm{B}}$. We have two situations for each $H_i^\prime \subseteq \bm{H}^\prime$.
\begin{enumerate}
    \item In $\widebar{\mathbb{M}}$, there does not exist an undirected path between $H_i^\prime$ and $\bm{A}^\prime$ or an undirected path between $H_i^\prime$ and $\bm{B}^\prime$ (or both) that is active given $\bm{C}^\prime$. More specifically:
    \begin{enumerate}
    \item If there does not exist an undirected path between $H_i^\prime$ and $\bm{A}^\prime$ that is active given $\bm{C}^\prime$, but such a path exists between $H_i^\prime$ and $\bm{B}^\prime$, then include $H_i^\prime$ into $\ddot{\bm{B}}$.
    \item If there does not exist an undirected path between $H_i^\prime$ and $\bm{B}^\prime$ that is active given $\bm{C}^\prime$, but such a path exists between $H_i^\prime$ and $\bm{A}^\prime$, then include $H_i^\prime$ into $\ddot{\bm{A}}$.
    \item If there does not exist an undirected path between $H_i^\prime$ and $\bm{A}^\prime$ that is active given $\bm{C}^\prime$ and there likewise does not exist such a path between $H_i^\prime$ and $\bm{B}^\prime$, then include $H_i^\prime$ into $\ddot{\bm{A}}$.
    \end{enumerate}
    \item In $\widebar{\mathbb{M}}$, there exists an undirected path between $H_i^\prime$ and $\bm{A}^\prime$ and an undirected path between $H_i^\prime$ and $\bm{B}^\prime$ that are both active given $\bm{C}^\prime$. We have two cases:
    \begin{enumerate}
    \item There exists an undirected path between $H_i^m$ and $\bm{A}^\prime$ and an undirected path between $H_i^m$ and $\bm{B}^\prime$ that are both active given $\bm{C}^\prime$. But this implies that $\bm{A}^\prime$ and $\bm{B}^\prime$ are connected given $\bm{C}^\prime$ in $\widebar{\mathbb{M}}$ via $H_i^m$ - a contradiction. 
    \item There exists an undirected path between $H_i^m$ and $\bm{A}^\prime$ and an undirected path between $H_i^n$ ($m \not = n)$ and $\bm{B}^\prime$ that are both active given $\bm{C}^\prime$ - denote these paths by $\Pi_{H_i^m \bm{A}^\prime}$ and $\Pi_{H_i^n \bm{B}^\prime}$, respectively. Note that there does \textit{not} exist an undirected path between $H_i^n$ and $\bm{A}^\prime$ and an undirected path between $H_i^m$ and $\bm{B}^\prime$ that are both active given $\bm{C}^\prime$ per the argument in (a). We have two cases:
    \begin{enumerate}
        \item There does not exist a descendant of $\bm{T}$ on $\Pi_{H_i^m \bm{A}^\prime}$. Then $\Pi_{H_i^m \bm{A}^\prime}$ must be confined to the sub-graph $\mathbb{G}^m$ in $\widebar{\mathbb{M}}$. But then an analogous undirected path $\Pi_{H_i^n \bm{A}^\prime}$ must be active in the sub-graph $\mathbb{G}^n$ - a contradiction. A similar argument also applies to $\Pi_{H_i^n \bm{B}^\prime}$. 
        \item Let $\bm{R}$ denote all members of $\bm{T}$ that have a descendant on $\Pi_{H_i^m \bm{A}^\prime}$. We will construct an active path between $H_i^n$ and $\bm{A}^\prime$ in $\widebar{\mathbb{M}}$ for a contradiction. We construct a path from $H_i^n$ in $\mathbb{G}^n$ using the corresponding vertices on $\Pi_{H_i^m \bm{A}^\prime}$ in $\mathbb{G}^m$ until we encounter the first child of $\bm{R}$, denoted by $X_j^n$. Consider the collection $\mathcal{P} = \{ \Pi_{H_i^n X_j^n}, X_j^n - R_k - X_j^m, \Pi_{X_j^m \bm{A}^\prime} \}$, for some $R_k \in \bm{R}$, whose concatenation connects $H_i^n$ and $\bm{A}^\prime$ given $\bm{C}^\prime$ in $\widebar{\mathbb{M}}$ - a contradiction. A similar argument also applies to $\Pi_{H_i^n \bm{B}^\prime}$.
        \end{enumerate}
    \end{enumerate}
    We have exhausted all possibilities and therefore conclude that there cannot exist an undirected path between $H_i^\prime$ and $\bm{A}^\prime$ and an undirected path between $H_i^\prime$ and $\bm{B}^\prime$ that are both active given $\bm{C}^\prime$.
\end{enumerate}
We have constructed a disjoint partition of vertices $(\ddot{\bm{A}}, \ddot{\bm{B}})$ such that $\ddot{\bm{A}} \cup \ddot{\bm{B}} \cup \bm{C}^\prime = \widebar{\textnormal{Anc}}_{\mathbb{M}}(\bm{A}^\prime\cup\bm{B}^\prime \cup \bm{C}^\prime ) $. Moreover, $\ddot{\bm{A}}$ and $\ddot{\bm{B}}$ are separated given $\bm{C}^\prime$ in $\widebar{\mathbb{M}}$ because, if an active path exists between $\ddot{\bm{A}}$ and $\ddot{\bm{B}}$ given $\bm{C}^\prime$, this implies the contradiction that there also exists an active path between $\ddot{\bm{A}}$ and $\bm{B}^\prime$ given $\bm{C}^\prime$.

We may then consider all of the cliques in $\widebar{\mathbb{M}}$ corresponding to each vertex and its married parents. Denote this set of cliques as $\mathcal{E}$. Also let $\mathcal{E}_{\ddot{\bm{B}}}$ denote the set of cliques in $\mathcal{E}$ that have non-empty intersection with $\ddot{\bm{B}}$. Because $\ddot{\bm{A}}$ and $\ddot{\bm{B}}$ are separated given $\bm{C}^\prime$, the vertices $\ddot{\bm{A}}$ and $\ddot{\bm{B}}$ are also non-adjacent in $\widebar{\mathbb{M}}$; this implies that no clique in $\mathcal{E}_{\ddot{\bm{B}}}$ can contain a member of $\ddot{\bm{A}}$. We also have $\ddot{\bm{B}} \cap e =\emptyset$ for all $e \in \mathcal{E} \setminus \mathcal{E}_{\ddot{\bm{B}}}$.

Consider an arbitrary graph $\mathbb{G}^j \in \mathcal{G}$. We can write:
\begin{equation} \nonumber
\begin{aligned}
   f^j(\ddot{\bm{A}}, \ddot{\bm{B}}, \bm{C}) &=  \prod_{\{D_i \cup \textnormal{Pa}^j(D_i)\} \in \mathcal{E}^j \setminus \mathcal{E}^j_{\ddot{\bm{B}}}} f(D_i | \textnormal{Pa}^j(D_i) ) \prod_{\{D_i \cup \textnormal{Pa}^j(D_i)\} \in  \mathcal{E}^j_{\ddot{\bm{B}}}} f(D_i | \textnormal{Pa}^j(D_i) )\\
    &=\prod_{e \in \mathcal{E}^j \setminus \mathcal{E}^j_{\ddot{\bm{B}}}} \gamma^j(e) \prod_{e \in \mathcal{E}^j_{\ddot{\bm{B}}}} \gamma^j(e) = \gamma^j(\ddot{\bm{A}}, \bm{C} )  \gamma^j(\ddot{\bm{B}}, \bm{C} ),
\end{aligned}
\end{equation}
where $\gamma^j$ is a placeholder for some non-negative function for $\mathbb{G}^j$. Let $\bm{U} = \bm{T} \cap (\ddot{\bm{A}} \cup \bm{C})$ and $\bm{V} = \bm{T} \cap (\ddot{\bm{B}} \cup \bm{C})$. Also let $\mathcal{U}$ denote the sub-sets of the sets in $\mathcal{T}$ corresponding to $\bm{U}$ - likewise for $\mathcal{V}$. We then proceed by integrating out $[\ddot{\bm{A}} \cup\ddot{\bm{B}}] \setminus [\bm{A}\cup\bm{B}]$:
\begin{equation} \nonumber
\begin{aligned}
f(\bm{A}, \bm{B}, \bm{C}) &=\sum_{[\ddot{\bm{A}}\cup\ddot{\bm{B}}] \setminus [\bm{A}\cup\bm{B}]} f(\ddot{\bm{A}}, \ddot{\bm{B}}, \bm{C})\\
&=\sum_{[\ddot{\bm{A}}\cup\ddot{\bm{B}}] \setminus [\bm{A}\cup\bm{B}]} \sum_{j=1}^q \mathbbm{1}_{\bm{T} \in \mathcal{T}^j}
f^j(\ddot{\bm{A}}, \ddot{\bm{B}}, \bm{C}) \\
&=\sum_{[\ddot{\bm{A}}\cup\ddot{\bm{B}}] \setminus [\bm{A}\cup\bm{B}]} \sum_{j=1}^q \mathbbm{1}_{\bm{U} \in \mathcal{U}^j} \mathbbm{1}_{\bm{V} \in \mathcal{V}^j} f^j(\ddot{\bm{A}}, \ddot{\bm{B}}, \bm{C})\\
&=\sum_{[\ddot{\bm{A}}\cup\ddot{\bm{B}}] \setminus [\bm{A}\cup\bm{B}]}
\gamma(\ddot{\bm{A}}, \bm{C} )  \gamma(\ddot{\bm{B}}, \bm{C} ),
\end{aligned}
\end{equation}
where $\gamma(\ddot{\bm{A}},\bm{C}) = \gamma^j(\ddot{\bm{A}},\bm{C})$ when $\bm{U} \in \mathcal{U}^j$ and $\gamma(\ddot{\bm{B}},\bm{C}) = \gamma^j(\ddot{\bm{B}},\bm{C})$ when $\bm{V} \in \mathcal{V}^j$.  We then finalize the integration:
\begin{equation} \nonumber
\begin{aligned}
&\hspace{5mm} \sum_{[\ddot{\bm{A}}\cup\ddot{\bm{B}}] \setminus [\bm{A}\cup\bm{B}]}
\gamma(\ddot{\bm{A}}, \bm{C} )  \gamma(\ddot{\bm{B}}, \bm{C} ) \\
&= \sum_{[\ddot{\bm{A}} \setminus \bm{A}] \cup [\ddot{\bm{B}} \setminus \bm{B}]} \gamma(\ddot{\bm{A}}, \bm{C} ) \gamma(\ddot{\bm{B}}, \bm{C})\\
&= \Big[\sum_{[\ddot{\bm{B}} \setminus \bm{B}]} \Big[ \sum_{[\ddot{\bm{A}} \setminus \bm{A}]} \gamma(\ddot{\bm{A}}, \bm{C}) \Big] \gamma(\ddot{\bm{B}}, \bm{C})\Big]\\
&= \sum_{[\ddot{\bm{A}} \setminus \bm{A}]} \gamma(\ddot{\bm{A}}, \bm{C})  \sum_{[\ddot{\bm{B}} \setminus \bm{B}]} \gamma(\ddot{\bm{B}}, \bm{C}) \\
&= \gamma(\bm{A}, \bm{C}) \gamma(\bm{B}, \bm{C}),
\end{aligned}
\end{equation}
The third equality follows because $[\ddot{\bm{A}} \setminus \bm{A}] \cap [\ddot{\bm{B}} \setminus \bm{B}]=\emptyset$ by construction. The conclusion follows by the last equality.
\end{proof}

\begin{replemma}{lem_anc1}
Suppose $O_i^\prime \ci_d O_j^\prime|\bm{W}^\prime \cup \bm{S}^\prime$ in $\mathbb{M}$ but $O_i^\prime \not \ci_d O_j^\prime|\bm{V}^\prime \cup \bm{S}^\prime$ for every $\bm{V} \subset \bm{W}$. If $O_k \in \bm{W}$, then $O_k \in \textnormal{Anc}_{\mathbb{F}}(\{O_i, O_j\} \cup \bm{S})$.
\end{replemma}
\begin{proof}
We invoke Lemma 15 in \citep{Strobl18} by setting $\bm{R}=\emptyset$, $O_i = O_i^\prime$, $O_j = O_j^\prime$, $\bm{W} = \bm{W}^\prime$ and $\bm{S} = \bm{S}^\prime$ in that paper. We can then conclude that $O_k^\prime \in \textnormal{Anc}_{\mathbb{M}}(O_i^\prime \cup O_j^\prime \cup \bm{S}^\prime)$. Moreover, if $O_k^\prime \in \textnormal{Anc}_{\mathbb{M}}(O_i^\prime \cup O_j^\prime \cup \bm{S}^\prime)$, then $O_k \in \textnormal{Anc}_{\mathbb{F}}(O_i \cup O_j \cup \bm{S})$ by construction of $\mathbb{F}$.
\end{proof}

\begin{reptheorem}{thm_sound}
Suppose the longitudinal density $f(\cup_{k=1}^w\tensor[^k]{\bm{O}}{}, \bm{L}, \bm{S})$ factorizes according to Equation \eqref{eq_fac_endM}. Assume that all arrowheads deduced from $\mathcal{P}$ are correct. Then, under d-separation faithfulness w.r.t. $\mathbb{M}$, the CIM algorithm returns the mixed graph $\mathbb{F}^*$ partially oriented.
\end{reptheorem}
\begin{proof}
Under d-separation faithfulness w.r.t. $\mathbb{M}$, CI and d-separation w.r.t. $\mathbb{M}$ are equivalent by Theorem \ref{thm_DMP}, so we can refer to them interchangeably. Algorithm \ref{pc_skel} finds the adjacencies in List \ref{list_adj} because we must always have $\tensor*[^{a}_b]{\textnormal{Adj}}{}_{\mathbb{F}^*}(\tensor[^a]{O}{_i}) \subseteq \tensor*[^{a}_b]{\textnormal{Adj}}{}_{\widehat{\mathbb{F}}^*}(\tensor[^a]{O}{_i})$ in Step \ref{pc_skel:new_set} of Algorithm \ref{pc_skel}. Step \ref{alg_tail1} discovers the correct tails by Lemma \ref{lem_anc1}. Step \ref{alg_tail2} follows directly by transitivity of the tails. \end{proof}

\subsection{Skeleton Discovery}\label{app_skel}

We summarize CIM's skeleton discovery procedure in Algorithm \ref{pc_skel}. The algorithm mimics PC-stable's skeleton discovery procedure, but it incorporates wave information in the adjacency sets.
\begin{algorithm}[]
 \KwData{CI oracle, $\mathcal{W}$}
 \KwResult{$\widehat{\mathbb{F}}^*$, \textnormal{Sep}}
 \BlankLine
 
 Form a complete graph $\widehat{\mathbb{F}}^*$ over $\bm{O}$ with edges $\circlinecirc$ \label{pc_skel:full}\\
 $l \leftarrow -1$ \\
 \Repeat{all pairs of adjacent vertices $(\tensor[^a]{O}{_i}, \tensor[^b]{O}{_j})$ in $\widehat{\mathbb{F}}^*$ satisfy $|\tensor*[^a_b]{\textnormal{Adj}}{_{\widehat{\mathbb{F}}^*}}(\tensor[^a]{O}{_i})\setminus \tensor[^b]{O}{_j}| \leq l$}{
 $l=l+1$ \\
 \Repeat{all ordered pairs of adjacent vertices $(\tensor[^a]{O}{_i}, \tensor[^b]{O}{_j})$ in $\widehat{\mathbb{F}}^*$ with $|\tensor*[^a_b]{\textnormal{Adj}}{_{\widehat{\mathbb{F}}^*}}(\tensor[^a]{O}{_i})\setminus \tensor[^b]{O}{_j}| \geq l$ have been considered}{
 Select a new ordered pair of vertices $(\tensor[^a]{O}{_i}, \tensor[^b]{O}{_j})$ that are adjacent in $\widehat{\mathbb{F}}^*$ and satisfy $|\tensor*[^a_b]{\textnormal{Adj}}{_{\widehat{\mathbb{F}}^*}}(\tensor[^a]{O}{_i})\setminus \tensor[^b]{O}{_j}| \geq l$ \\
 
 \Repeat{$\tensor[^a]{O}{_i}$ and $\tensor[^b]{O}{_j}$ are no longer adjacent in $\widehat{\mathbb{F}}^*$ or all $\bm{W} \subseteq \tensor*[^a_b]{\textnormal{Adj}}{_{\widehat{\mathbb{F}}^*}}(\tensor[^a]{O}{_i})\setminus \tensor[^b]{O}{_j}$ with $|\bm{W}| = l$ have been considered }{
 
 Choose a new set $\bm{W} \subseteq \tensor*[^a_b]{\textnormal{Adj}}{_{\widehat{\mathbb{F}}^*}}(\tensor[^a]{O}{_i})\setminus \tensor[^b]{O}{_j}$ with $|\bm{W}|=l$ \label{pc_skel:new_set}\\
 
 \If{$\tensor[^a]{O}{_i} \ci \tensor[^b]{O}{_j}|\bm{W} \cup \bm{S}$}{ \label{pc_skel:query}
 	Delete the edge $\tensor[^a]{O}{}_i \circlinecirc \tensor[^b]{O}{}_j$ from $\widehat{\mathbb{F}}^*$ \\
 	$\textnormal{Sep}(\tensor[^a]{O}{}_i,\tensor[^b]{O}{}_j) \leftarrow \textnormal{Sep}(\tensor[^b]{O}{}_j,\tensor[^a]{O}{}_i) \leftarrow \bm{W}$
 }

 }
 }
 }

 \BlankLine

 \caption{CIM's skeleton discovery procedure} \label{pc_skel}
\end{algorithm}

\subsection{Comparison to Previous Global Markov Property}
\label{sec_spirtes}
\citet{Spirtes94} also characterized the global Markov property across a mixture of DAGs using $\mathbb{F}$. The fused graph however implies less CI relations than $\mathbb{M}$ as illustrated in Figure \ref{fig_MGoo}. We have drawn $\mathbb{F}$ in Figure \ref{fig_MGcc}. $X_1$ and $X_3$ are d-connected in $\mathbb{F}$ even though $\{X_1^1, X_1^2\}$ and $\{X_3^1, X_3^2\}$ are d-separated in $\mathbb{M}$ in Figure \ref{fig_MGbb}. We have established an instance where the mixture graph implies strictly more independence relations than the fused graph.

The mixture graph in fact always implies at least the same number of CI relations as the fused graph:
\begin{proposition1} \label{prop_F_M}
Let $\bm{A},\bm{B},\bm{C}$ denote disjoint subsets of $\bm{X}$. If $\bm{A} \ci_d \bm{B} | \bm{C}$ in $\mathbb{F}$, then $\bm{A}^\prime \ci_d \bm{B}^\prime | \bm{C}^\prime$ in $\mathbb{M}$.
\end{proposition1}
\begin{proof}
We create $q$ copies of $\mathbb{F}$ and plot them adjacent to each other. Denote the resultant graph as $\mathbb{F}^\prime$. As a result, we have $\bm{A} \ci_d \bm{B} | \bm{C}$ in $\mathbb{F}$ if and only if $\bm{A}^\prime \ci_d \bm{B}^\prime | \bm{C}^\prime$ in $\mathbb{F}^\prime$. Create a new graph $\mathbb{F}^{\prime \prime}$ as follows. First set $\mathbb{F}^{\prime \prime}$ equal to $\mathbb{F}^\prime$. Then remove $\bm{T}^\prime$ from $\mathbb{F}^{\prime \prime}$ and place $\bm{T}$ instead. Set $\textnormal{Ch}_{\mathbb{F}^{\prime\prime}}(T_i)$ equal to $\textnormal{Ch}^\prime_{\mathbb{F}}(T_i)$ for each $T_i \in \bm{T}$. Denote the resultant graph as $\mathbb{F}^{\prime \prime}$.

We will show that $\bm{A}^\prime \not \ci_d \bm{B}^\prime | \bm{C}^\prime$ in $\mathbb{F}^{\prime \prime}$ implies $\bm{A}^\prime \not \ci_d \bm{B}^\prime | \bm{C}^\prime$ in $\mathbb{F}^\prime$. Consider any active path $\Pi_{\bm{A}^\prime \bm{B}^\prime}$ between $\bm{A}^\prime$ and $\bm{B}^\prime$ given $\bm{C}^\prime$ in $\mathbb{F}^{\prime \prime}$. Denote the moral graph of $\textnormal{Anc}_{\mathbb{F}^{\prime \prime}}(\bm{A}^\prime,\bm{B}^\prime, \bm{C}^\prime)$ by $\widebar{\mathbb{F}}^{\prime \prime}$. Consider an active path $\Pi_{\bm{A}^\prime \bm{B}^\prime}$ between $\bm{A}^\prime$ and $\bm{B}^\prime$ given $\bm{C}^\prime$ in $\widebar{\mathbb{F}}^{\prime \prime}$. We can replace an arbitrary vertex $Z_i^m$ on $\Pi_{\bm{A}^\prime \bm{B}^\prime}$ with $Z_i^n$ on $\mathbb{G}^n \in \mathcal{G}$. Repeating this process for every vertex on $\Pi_{\bm{A}^\prime \bm{B}^\prime}$ creates a non-simple path (i.e. with potentially repeated vertices) between $\bm{A}^n$ and $\bm{B}^n$ that does not contain any member of $\bm{C}^n$. There thus exists a simple path without repeated vertices between $\bm{A}^n$ and $\bm{B}^n$ that does not contain any member of $\bm{C}^n$ in $\widebar{\mathbb{F}}^{\prime \prime}$. Hence $\bm{A}^\prime \not \ci_d \bm{B}^\prime | \bm{C}^\prime$ in $\mathbb{F}^\prime$ by Lemma 2 in \citep{Lauritzen90}.

Note that all of the edges in $\mathbb{M}$ are contained within $\mathbb{F}^{\prime \prime}$. The conclusion follows because we may write $\bm{A}^\prime \not \ci_d \bm{B}^\prime | \bm{C}^\prime$ in $\mathbb{M}$ implies $\bm{A}^\prime \not \ci_d \bm{B}^\prime | \bm{C}^\prime$ in $\mathbb{F}^{\prime \prime}$, which implies $\bm{A}^\prime \not \ci_d \bm{B}^\prime | \bm{C}^\prime$ in $\mathbb{F}^\prime$, which implies $\bm{A} \not \ci_d \bm{B} | \bm{C}$ in $\mathbb{F}$.

\end{proof}
\noindent $\mathbb{M}$ is thus superior to $\mathbb{F}$ because (1) $\mathbb{M}$ implies at least as many CI relations as $\mathbb{F}$, and (2) $\mathbb{M}$ implies strictly more CI relations in some cases.

\subsection{Negative Result} \label{app_imp}
We cannot infer arrowheads with a CI oracle alone:
\begin{proposition1}
There exist mother and fused graph pairs $(\mathbb{M}_1,\mathbb{F}_1)$ and $(\mathbb{M}_2,\mathbb{F}_2)$ such that $O_i \not \in \textnormal{Anc}_{\mathbb{F}_1}(O_j \cup \bm{S})$ and $O_i \in \textnormal{Anc}_{\mathbb{F}_2}(O_j)$, but $O_i^\prime \ci O_j^\prime | (\bm{W}^\prime,\bm{S}^\prime)$ in $\mathbb{M}_1$ if and only if $O_i^\prime \ci O_j^\prime | (\bm{W}^\prime,\bm{S}^\prime)$ in $\mathbb{M}_2$ for any $O_i,O_j \in \bm{O}$ and $\bm{W} \subseteq \bm{O} \setminus \{O_i,O_j\}$.
\end{proposition1}
\begin{proof}
Assume $O_i \not \in \textnormal{Anc}_{\mathbb{F}_1}(O_j \cup \bm{S})$. Let $\mathcal{G}_1$ and $\mathcal{G}_2$ refer to the set of DAGs associated with $\mathbb{M}_1$ and $\mathbb{M}_2$, respectively. Choose $\mathbb{M}_1$ arbitrarily and set $\mathcal{G}_2$ equal to $\mathcal{G}_1$. If $|\mathcal{G}_2|=1$, then add a second copy of the DAG into $\mathcal{G}_2$. Add one new latent variable $L_k$ into each DAG in $\mathcal{G}_2$ as follows. For all but the last DAG, draw the directed edge $O_i \rightarrow L_k$. For the last DAG, draw $L_k \rightarrow O_j$. Next, introduce a new latent common cause $T_l$ for $L_k$ and $O_j$ into every DAG in $\mathcal{G}_2$. The new paths do not introduce a d-connecting path between the observed vertices in any of the DAGs in $\mathcal{G}_2$. As a result, $O_i^\prime \ci O_j^\prime | (\bm{W}^\prime,\bm{S}^\prime)$ in $\mathbb{M}_1$ if and only if $O_i^\prime \ci O_j^\prime | (\bm{W}^\prime,\bm{S}^\prime)$ in $\mathbb{M}_2$, but $O_i \in \textnormal{Anc}_{\mathbb{F}_2}(O_j)$ with the directed path $O_i \rightarrow L_k \rightarrow O_j$.
\end{proof}

\subsection{Failure of Other Constraint-Based Methods} \label{app_fail}

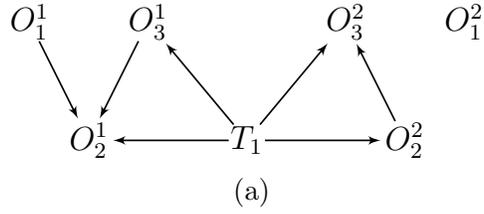
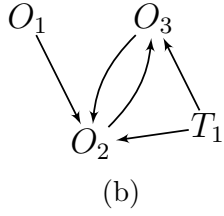
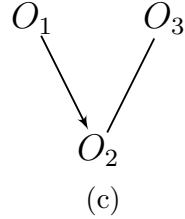
\begin{figure*}
\centering
\begin{subfigure}{0.42\textwidth}
\centering
\resizebox{\linewidth}{!}{
\begin{tikzpicture}[scale=1.0, shorten >=1pt,auto,node distance=2.8cm, semithick]
                    
\tikzset{vertex/.style = {inner sep=0.4pt}}
\tikzset{edge/.style = {->,> = latex'}}
 
\node[vertex] (1) at  (0,0) {$O_1^1$};
\node[vertex] (2) at  (0.75,-1.5) {$O_2^1$};
\node[vertex] (3) at  (1.5,0) {$O_3^1$};
\node[vertex] (7) at  (2.75,-1.5) {$T_1$};

\draw[edge] (1) to (2);
\draw[edge] (3) to (2);
\draw[edge] (7) to (2);
\draw[edge] (7) to (3);


\node[vertex] (3) at  (4,0) {$O_3^2$};
\node[vertex] (2) at  (4.75,-1.5) {$O_2^2$};
\node[vertex] (1) at  (5.5,0) {$O_1^2$};

\draw[edge] (2) to (3);
\draw[edge] (7) to (2);
\draw[edge] (7) to (3);
\end{tikzpicture}
}
\caption{}  \label{fig_FCI_fail:mother}
\end{subfigure}
\vspace{10mm}

\begin{subfigure}{0.205\textwidth}
\centering
\resizebox{\linewidth}{!}{
\begin{tikzpicture}[scale=1.0, shorten >=1pt,auto,node distance=2.8cm, semithick]
                    
\tikzset{vertex/.style = {inner sep=0.4pt}}
\tikzset{edge/.style = {->,> = latex'}}
 
\node[vertex] (1) at  (0,0) {$O_1$};
\node[vertex] (3) at  (1.5,0) {$O_3$};
\node[vertex] (2) at  (0.75,-1.5) {$O_2$};

\draw[edge] (1) to (2);
\draw[edge, bend right=20] (2) to (3);
\draw[edge, bend right=20] (3) to (2);

\node[vertex] (7) at  (2.15,-1.3) {$T_1$};
\draw[edge] (7) to (2);
\draw[edge] (7) to (3);

\end{tikzpicture}
}
\caption{}  \label{fig_FCI_fail:summary}
\end{subfigure}
\hspace{10mm}
\begin{subfigure}{0.17\textwidth}
\centering
\resizebox{\linewidth}{!}{
\begin{tikzpicture}[scale=1.0, shorten >=1pt,auto,node distance=2.8cm, semithick]
                    
\tikzset{vertex/.style = {inner sep=0.4pt}}
\tikzset{edge/.style = {->,> = latex'}}
 
\node[vertex] (1) at  (0,0) {$O_1$};
\node[vertex] (3) at  (1.5,0) {$O_3$};
\node[vertex] (2) at  (0.75,-1.5) {$O_2$};

\draw[edge] (1) to (2);
\draw[] (2) to (3);

\end{tikzpicture}
}
\caption{}  \label{fig_FCI_fail:summary_MG}
\end{subfigure}

\caption{An example where both FCI and CCI fail. We have a mixture graph in (a) and its fused graph in (b). Subfigure (c) contains the correct $\mathbb{F}^*$, but FCI and CCI infer the incorrect collider $O_1 * \!\! \rightarrow O_2 \leftarrow \!\! * O_3$.}\label{fig_FCI_fail}
\end{figure*}

We cannot apply an existing constraint-based algorithm on data arising from a mixture of DAGs and recover a partially oriented $\mathbb{F}^*$. For example, FCI and CCI can make incorrect inferences if $\mathcal{G}$ contains more than one DAG. Consider the mixture graph in Figure \ref{fig_FCI_fail:mother}, where all variables lie in the same wave. $O_2$ is an ancestor of $O_3$ in $\mathbb{F}$ drawn in Figure \ref{fig_FCI_fail:summary}, but we have $\{O_1^1,O_1^2\}$ $\ci_d \{O_3^1,O_3^2\}$ in $\mathbb{M}$, so $O_1$ and $O_3$ are independent by Theorem \ref{thm_DMP}. FCI and CCI therefore infer the incorrect collider $O_1 * \!\! \rightarrow O_2 \leftarrow \!\! * O_3$ in $\mathbb{F}^*$ during v-structure discovery. We thus require an alternative algorithm to correctly recover a partially oriented $\mathbb{F}^*$.

\end{document}